\documentclass{article}

\usepackage[final,nonatbib]{neurips_2022}

\usepackage{graphicx}
\usepackage{tikz}
\usepackage{comment}
\usepackage{amsmath,amssymb} % define this before the line numbering.
\usepackage{color}
\usepackage[accsupp]{axessibility}  % Improves PDF readability for those with disabilities.

\newcommand{\real}{\mathbb{R}}

\usepackage[utf8]{inputenc} % allow utf-8 input
\usepackage[T1]{fontenc}    % use 8-bit T1 fonts
\usepackage{hyperref}       % hyperlinks
\hypersetup{
    colorlinks=true,
    linkcolor=red,
    urlcolor=magenta,
    pdfpagemode=FullScreen,
}
\usepackage{url}            % simple URL typesetting
\usepackage{booktabs}       % professional-quality tables
\usepackage{amsfonts}       % blackboard math symbols
\usepackage{nicefrac}       % compact symbols for 1/2, etc.
\usepackage{microtype}      % microtypography
\usepackage{xcolor}         % colors
\usepackage[square,numbers]{natbib}

\usepackage{gensymb}
\usepackage{booktabs}
\usepackage{animate}

\usepackage{caption}
\usepackage{tabu}
\usepackage{multirow,bigstrut}
\usepackage{cancel}
\usepackage{tabularx}
\usepackage{placeins}
\usepackage{float}

\usepackage{url}

\newcommand{\centered}[1]{\begin{tabular}{@{}l@{}} #1 \end{tabular}}

\title{Controllable Video Generation through\\ Global and Local Motion Dynamics}

\author{%
  Aram Davtyan \\
  Computer Vision Group \\
  University of Bern, Switzerland\\
  \texttt{aram.davtyan@inf.unibe.ch} \\
  \And
  Paolo Favaro \\
  Computer Vision Group \\
  University of Bern, Switzerland \\
  \texttt{paolo.favaro@inf.unibe.ch} \\
}

\begin{document}

\maketitle

\begin{abstract}
  We present GLASS, a method for Global and Local Action-driven Sequence Synthesis. GLASS is a generative model that is trained on video sequences in an unsupervised manner and that can animate an input image at test time. The method learns to segment frames into foreground-background layers and to generate transitions of the foregrounds over time through a global and local action representation. Global actions are explicitly related to 2D shifts, while local actions are instead related to (both geometric and photometric) local deformations. GLASS uses a recurrent neural network to transition between frames and is trained through a reconstruction loss. We also introduce W-Sprites (Walking Sprites), a novel synthetic dataset with a predefined action space. We evaluate our method on both W-Sprites and real datasets, and find that GLASS is able to generate realistic video sequences from a single input image and to successfully learn a more advanced action space than in prior work. %Further details, the code and example videos are available at \url{github}.
\end{abstract}

\section{Introduction}

A long-standing objective in machine learning and computer vision is to build agents that can learn how to operate in an environment through visual data \cite{finn2016unsupervised}. A successful approach to do so is to use supervised learning, \emph{i.e.}, to train a model on a large, manually annotated dataset \cite{mottaghi2016newtonian}. However, if we take inspiration from how infants learn to move, we are brought to conclude that they may not rely on extensive guidance. In fact, while supervision from adults might come through language \cite{smith2005development}, the signal is certainly not detailed enough to fully define the locomotion dynamics. 
One approach that does not require direct supervision is to learn just through direct scrutiny of other agents, \emph{i.e.}, through passive imitation. In fact, infants have an abundance of sensory exposure to the activities of adults before they themselves learn how to perform them \cite{rybkin2018learning}.

The first step for an observing agent to learn how to operate in an environment through passive imitation and without explicit supervision is to build a model that: 
%Inspired by these observations, we propose a method to learn how agents move directly from video sequences and without any manually defined supervision signal. In our view, the model of an observing agent should: 
1) separates an agent from its environment, 2) captures the appearance of the agent and its environment, and 3) builds a description of the agent's dynamics. The first requirement implies that the model incorporates some segmentation capability, and it allows to explain transitions over time more easily. The second requirement is dictated by the fact that we exploit the reconstruction of visual observations as our indirect supervision signal. Thus, our model also relates to the video generation literature. Finally, the third requirement is that the model includes an \emph{action space}, which serves two purposes: i) it allows the model to decode a video into a sequence of actions (which is a representation of the agent's dynamics) and ii) it allows the model to control the generation of videos by editing the action sequence.

Recently, several methods have explored the above direction to different extents \cite{menapace2021playable,huang2021layered,hu2021learning,rybkin2018learning,smirnov2021marionette}. In particular, CADDY \cite{menapace2021playable} models the dynamics and the appearance of videos globally: It does not have an explicit separation between the agent and the environment, and thus their dynamics are tangled in action space (\emph{e.g.}, a person walking and the camera panning). 
A recent method by Huang et~al.~\cite{huang2021layered} is instead heavily based on the segmentation of a foreground from the background. However, this method does not learn to decode a sequence of frames into an action sequence. Moreover, the method relies on a static background and  the dynamics are limited to transformations of a mask (2D shifts and affine transformations) so that it would not be obvious how to control more general dynamics such as rotations and similar in situ animations.

\begin{figure}[t]
    \centering
    \animategraphics[width=0.24\textwidth]{7}{Figures/wsprites/ws_01_0}{1}{11}
    \animategraphics[width=0.24\textwidth]{7}{Figures/wsprites/ws_02_0}{1}{11}
    \animategraphics[width=0.24\textwidth]{7}{Figures/wsprites/ws_03_0}{2}{12}
    \animategraphics[width=0.24\textwidth]{7}{Figures/wsprites/ws_04_0}{2}{12}
    \caption{\emph{W-Sprites} dataset sample videos. To play them use Acrobat Reader.}
    \label{fig:wsprites}
\end{figure}

\begin{figure}[t]
    \centering
    \includegraphics[width=\textwidth,trim=0 0 0 19cm,clip]{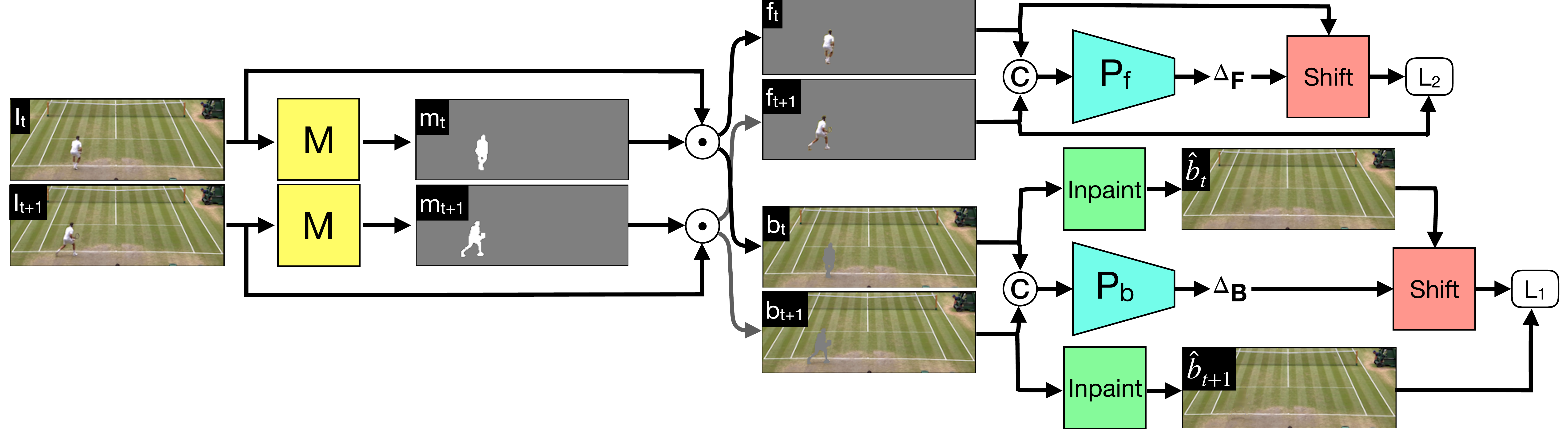}
    \caption{GLASS Global Motion Analysis. Two input frames $I_t$ and $I_{t+1}$ are fed (separately) to a segmentation network $M$ to output the foreground masks $m_t$ and $m_{t+1}$ respectively. The masks are used to separate the foregrounds $f_t$ and $f_{t+1}$ from the backgrounds $b_t$ and $b_{t+1}$. The concatenated foregrounds are fed to the network $P_f$ to predict their relative shift $\Delta_F$. We use $\Delta_F$ to shift $f_t$ and match it to $f_{t+1}$ via an $L_2$ loss (foregrounds may not match exactly and this loss does not penalize small errors). In the case of the backgrounds we also train an inpainting network before we shift them with the predicted $\Delta_B$ and match them with an $L_1$ loss (unlike foregrounds, we can expect backgrounds to match).
    \label{fig:GLASS_GMA}}
\end{figure}

\begin{figure}[t]
    \centering
    \includegraphics[width=\textwidth,trim=0 21.9cm 0 1.05cm, clip]{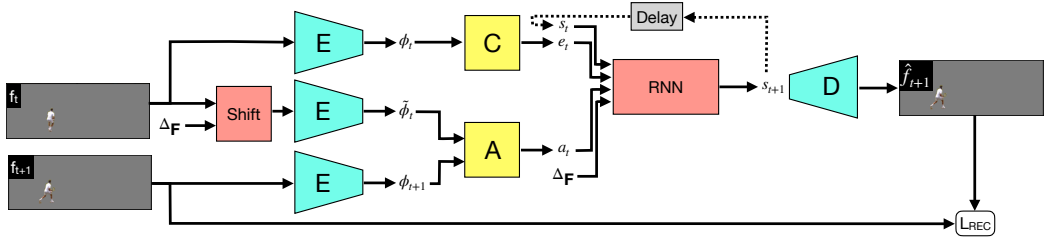}
    \caption{GLASS Local Motion Analysis. We feed the segmented foreground $f_t$, its shifted version and $f_{t+1}$ separately as inputs to an encoder network $E$ to obtain features $\phi_t$, $\tilde\phi_t$ and $\phi_{t+1}$ respectively. The latter two features are then mapped to an action $a_t$ by the action network $A$. A further encoding of $\phi_t$ into $e_t$, the previous state $s_t$, and the local action $a_t$ and global action $\Delta_F$ are fed as input to the RNN to predict the next state $s_{t+1}$. Finally, a decoder maps the state $s_{t+1}$ to the next foreground $\hat f_{t+1}$, which is matched to the original foreground $f_{t+1}$ via the reconstruction loss. %There is no gradient feeding back from the local motion networks to the global motion ones.
    \label{fig:GLASS_LMA}}
\end{figure}

To address these limitations, we introduce GLASS, a method for Global and Local Action-driven Sequence Synthesis. As shown in Fig.~\ref{fig:GLASS_GMA}, GLASS first learns to segment each frame of a video into foreground and background layers. A basic principle to do that is to use motion as a cue, \emph{i.e.}, the fact that agents exhibit, on average, a distinct motion flow compared to the environment. Motion-based segmentation could be achieved through background subtraction, which is however restricted to stationary backgrounds, or instead, more in general, via optical flow. For simplicity, we propose to use an explicit foreground-background motion segmentation based on 2D shifts. 
Then, GLASS regresses the relative shift between the foregrounds of two subsequent frames, which we call the \emph{global action}, and between the backgrounds (see Fig.~\ref{fig:GLASS_LMA}). The local actions are learned only from the foregrounds. We train an RNN to predict, through a decoder, the next foreground by using an encoding of a foreground, the previous state, and an encoding of the local and global actions as input. All networks are trained via reconstruction losses. 

We evaluate GLASS on both synthetic and real data. As synthetic data we introduce \emph{W-Sprites} (Walking Sprites \cite{li2018disentangle,liberatedpixelcup,universallpcspritesheet}) (see Fig.~\ref{fig:wsprites}), a dataset with a pre-defined action space, and where the action labels between pairs of frames (as well as the agent segmentation and location, and the background shift) are known. We find that GLASS learns a robust representation of both global and local dynamics on W-Sprites. Moreover, GLASS is able to decode videos into sequences of actions that strongly correlate with the ground truth action sequences. Finally, users can generate novel sequences by controlling the input action sequences to GLASS. On real data, we find that GLASS can also generate realistic sequences by controlling the actions between frames.\\
\noindent\textbf{Contributions:} i) We introduce GLASS, a novel generative model with a global and local action space; the shifts estimated and generated through the global actions have an accuracy comparable to or higher than SotA; moreover, local actions allow a fine-grained modeling of dynamics that is not available in prior work;
ii) We introduce W-Sprites, a novel dataset for the evaluation of action identification and generation; %, where both background and agents shift independently, agents move according to a known action space and the segmentation masks of the agents are known; W-Sprites can be used to evaluate the action space and decoding capability of a method; 
iii) We demonstrate GLASS on both synthetic and real datasets and show that it can: 1) segment an agent from its environment and estimate its global shift over time; 2) learn a disentangled action space that is consistent across agents; 3) decode videos into sequences of actions; 4) synthesize realistic videos under the guidance of a novel action policy.

% \begin{itemize}
%     \item we are interested in learning how objects behave within an environment
    
%     \item a first step is to separate objects from their environment
    
%     \item a basic principle to do that is to use motion and the fact that objects exhibit, on average, a distinct motion flow compared to the environment
    
%     \item motion-based segmentation could be achieved through background subtraction, which is restricted to stationary backgrounds, or, more in general, optical flow
    
%     \item for simplicity, we propose to use an explicit foreground-background motion segmentation based on 2D shifts (rather than generic optical flows)
    
%     \item 

% \end{itemize}

\section{Prior work}

\noindent\textbf{Video generation.} Because GLASS is trained based on reconstruction losses, and it is built as a generative model, it relates to the generation of videos. Recent success in deep generative models for images \cite{dhariwal2021diffusion,karras2021alias,razavi2019generating} has aroused renewed interest in video generation. Several formulations tackling the problem of video generation exploit adversarial losses \cite{acharya2018towards,babaeizadeh2017stochastic,denton2018stochastic,lee2018stochastic,tulyakov2018mocogan,vondrick2015anticipating,vondrick2016generating,wang2020g3an}, autoregressive models \cite{weissenborn2019scaling} and use a wide range of network architectures from RNNs \cite{srivastava2015unsupervised} to transformers \cite{yan2021videogpt}.\\
\noindent\textbf{Controllable video generation.} In order to model the variety of possible motion continuations from a given image, one could condition the generation on an external signal. Existing methods could be categorized by the type of the driving signal, that varies from fine-grained control sequences, such as motion strokes \cite{hu2021learning} to more general ones, such as textual descriptions of the actions \cite{hu2021make}. Some approaches introduce structure into the latent space of the generative model by disentangling motion and appearance \cite{tulyakov2018mocogan,wang2020g3an}. This allows transfer of the motion from one video to another, which can also be considered as a type of conditioning.

Video generation models can also differ in how they apply conditioning. While some prior work uses per-video class labels \cite{kim2019unsupervised,wang2020imaginator}, \emph{e.g.}, actions performed in a short sequence of frames, others, as in GLASS, use conditioning at each step \cite{chiappa2017recurrent,finn2016unsupervised,kim2020learning,nunes2020action,oh2015action}. For instance, in \cite{finn2016unsupervised} the authors train a model to simulate the behavior of a robotic arm given the performed actions. Kim et al.~\cite{kim2020learning} introduce GameGAN,  a powerful generative model that can replace a game engine. It is trained to render the next frame given the current frame and the pressed keyboard action.
%Despite the high quality of the generated sequences, 
One limitation of these methods is that they require knowledge of the ground truth actions and hence are restricted to synthetic data, such as video games. To become applicable to real data, several recent methods that learn an action space of the agent from raw videos without fine-grained annotations have been proposed. For instance, Rybkin et al.~\cite{rybkin2018learning} propose a continuous latent space for the actions. They introduce arithmetical structure into their action space by exploiting the fact that two actions can be composed to get another action that would lead to the same result as when applying the original actions sequentially. In \cite{menapace2021playable} the continuous action space is replaced by a finite set. This allows a simpler control (playability) of the generated videos and favors interpretability of the learned actions. More recent work by Huang et al. \cite{huang2021layered} explicitly separates the foreground from the background and trains a network to predict the next frame given the current frame and the next segmentation mask. GLASS relates to this last family of methods as it also does not require any supervision signal.\\
% \cite{hu2021make} Make It Move: Controllable Image-to-Video Generation with Text Descriptions
% \cite{hu2021learning} Learning to take directions one step at a time
% \cite{kim2019unsupervised} Unsupervised keypoint learning for guiding class-conditional video prediction
% \cite{wang2020imaginator} Imaginator: Conditional spatio-temporal gan for video generation
% \cite{finn2016unsupervised} Unsupervised learning for physical interaction through video prediction
% \cite{chiappa2017recurrent} Recurrent environment simulators
% \cite{kim2020learning} Learning to simulate dynamic environments with gamegan
% \cite{nunes2020action} Action-conditioned benchmarking of robotic video prediction models: a comparative study
% \cite{oh2015action} Action-conditional video prediction using deep networks in atari games
% \cite{rybkin2018learning} Learning what you can do before doing anything; continuous actions
% \cite{menapace2021playable} Playable video generation; discrete actions
% \cite{huang2021layered} Layered Controllable Video Generation; masking + translations
\noindent\textbf{Unsupervised learning of structured representations.} In GLASS we propose to learn the global and local actions from video frames. While the global ones are defined as foreground 2D shifts, the local ones are represented as a discrete set of action codes. This leads to a latent clustering problem. In GLASS, we propose to solve it through variational inference \cite{kingma2013auto}. Some recent work learns structured representations from raw input data \cite{caron2018deep,burgess2018understanding}. The VQ-VAE \cite{van2017neural} formulation instead uses a discrete latent space and assumes a uniform distribution over the latent features. Recent advances in image and video generation has shown that such VQ-VAE based models have a remarkable performance \cite{razavi2019generating,yan2021videogpt} and this has encouraged us to adopt this approach.\\
% \cite{caron2018deep} Deep clustering for unsupervised learning of visual features
% \cite{kingma2013auto} Auto-encoding variational bayes
% \cite{burgess2018understanding} Understanding disentangling in $\beta $-VAE
\noindent\textbf{Unsupervised segmentation.} The high cost of annotation in segmentation datasets has motivated work for segmentation in an unsupervised way \cite{bielski2019emergence}. %Bielski and Favaro~\cite{bielski2019emergence} proposed to solve the unsupervised segmentation problem through a GAN formulation. The model generates the foreground and the background images along with the segmentation masks. The masks are perturbed and the composed image is fed to a discriminator network to favor meaningful segmentations. 
More recently, Smirnov et al.~\cite{smirnov2021marionette} decompose images into a background and a learnt dictionary of sprites.
However, when a dataset of videos is available, one can use the temporal correlation to obtain foreground-background layers. A traditional approach based on the zero or small background motion assumption is background subtraction \cite{bouwmans2014background,elgammal2000non,stauffer1999adaptive}. In \cite{alayrac2019controllable,alayrac2019visual} the authors explicitly construct multilayer videos by mixing two arbitrary sequences and train a network to separate them back. The trained model should be able to extract meaningful layers from the real videos as well. Vondrik et al.~\cite{vondrick2016generating} propose a generative model that synthesizes sequences by separately generating the foreground and background videos and by combining them with a mask. GLASS relates to these model, but, unlike \cite{vondrick2016generating} it allows explicit control of the foreground video dynamics. \\

% \cite{bielski2019emergence} Emergence of object segmentation in perturbed generative models

% \cite{wang2019learning} Learning unsupervised video object segmentation through visual attention

% \cite{araslanov2021dense} Dense Unsupervised Learning for Video Segmentation

% \cite{jojic2001learning} Learning flexible sprites in video layers

% \cite{pawan2008learning} Learning layered motion segmentations of video

% \cite{wang1994representing} Representing moving images with layers

% \cite{webster2009color} Color vision: Appearance is a many-layered thin

% \cite{achanta2012slic} SLIC superpixels compared to state-of-the-art superpixel methods

% \cite{alexe2010classcut} Classcut for unsupervised class segmentation

% \cite{hochbaum2009efficient} An efficient algorithm for co-segmentation

% \cite{bouwmans2014background} Background modeling and foreground detection for video surveillance

% \cite{elgammal2000non} Non-parametric model for background subtraction

% \cite{stauffer1999adaptive} Adaptive background mixture models for real-time tracking

% \cite{goyette2012changedetection} Changedetection. net: A new change detection benchmark dataset

% \cite{bielski2019emergence} Emergence of object segmentation in perturbed generative models

% \cite{smirnov2021marionette} Marionette: Self-supervised sprite learning

% \cite{alayrac2019visual} The visual centrifuge: Model-free layered video representations

% \cite{alayrac2019controllable} Controllable attention for structured layered video decomposition

% \cite{vondrick2016generating} Generating videos with scene dynamics

\section{Training GLASS}

%As shown in Figs.~\ref{fig:GLASS_GMA} and \ref{fig:GLASS_LMA}, 
GLASS consists of two stages: One is the Global Motion Analysis (GMA) (shown in Fig.~\ref{fig:GLASS_GMA}) and the other is the Local Motion Analysis (LMA) (shown in Fig.~\ref{fig:GLASS_LMA}). %We call the first stage Global Motion Analysis (GMA) and the second one Local Motion Analysis (LMA). 
GMA aims to separate the foreground agent from the background and to also regress the 2D shifts between foregrounds and backgrounds. LMA aims to learn a representation for local actions that can describe deformations other than 2D shifts. Towards this purpose it uses a Recurrent Neural Network (RNN) and a feature encoding of a frame and of the global and local actions as input.
Both GLA and LMA stages are jointly trained in an unsupervised manner. % by using several losses that we describe in the next sections.

\subsection{Global Motion Analysis}

Let us denote a video as a sequence of $T$ frames $I_t\in \real^{3\times H\times W}$, where $t=1,\dots,T$, and $3$, $H$ and $W$ denote the number of color channels, the height and the width of the frame. Although GLASS is trained with video sequences, we can illustrate all the training losses with a single pair $(I_t,I_{t+1})$ of frames. Each frame is fed to a mask network $\text{M}$ to output masks $m_t$ and $m_{t+1}$. The masks can take values between $0$ and $1$ (a sigmoid is used at the output), but are encouraged to take the extreme values through the following binarization loss
\begin{align}
  \textstyle  {\cal L}_\text{BIN} = \sum_t \min\{m_t,1-m_t\}.
\end{align}
We also discourage the mask from being empty or covering the whole frame by using a mask size loss
\begin{align}
   \textstyle {\cal L}_\text{SIZE} = \sum_t | \mathbb{E}[m_t] - \theta |,
\end{align}
where $\mathbb{E}[\cdot]$ denotes the average over all pixels and $\theta\in[0,1]$ is a tuning parameter (the percentage of image pixels covered by a mask on average). 
The masks are then used to extract the foregrounds $f_t = I_t \odot m_t$ and $f_{t+1} = I_{t+1} \odot m_{t+1}$ and the backgrounds $b_t = I_t \odot (1-m_t)$ and $b_{t+1} = I_{t+1} \odot (1-m_{t+1})$ ($\odot$ denotes the element-wise product). 
We assume that the foregrounds are approximately matching up to a relative shift $\bar \Delta_F$, \emph{i.e.}, that $f_{t+1}[p] \simeq \left(f_t \circ \bar \Delta_F\right)[p] \doteq f_t[p+\bar \Delta_F]$, for all pixel coordinates $p\in \Omega\subset \real^2$. 
We then concatenate the foregrounds and feed them as input to the pose network $P_f$ to regress the relative shift $\Delta_F=P_f([f_t,f_{t+1}])$ between $f_t$ and $f_{t+1}$. Since we do not have the ground truth shift $\bar \Delta_F$, we cannot train $P_f$ via supervised learning. In alternative, we rely on the modeling assumption and define a reconstruction loss for the foreground by applying the estimated shift $\Delta_F$ to $f_t$ and by matching it to the frame $f_{t+1}$ in the $L_2$ norm (to allow for some error tolerance), \emph{i.e.},
\begin{align}
    \textstyle {\cal L}_\text{RECF} = \sum_t \big\| f_{t+1} - f_t\circ \Delta_F \big\|_2^2.
\end{align}
A similar derivation pertains to the backgrounds. We concatenate the backgrounds and feed them as input to the pose network $P_b$ to regress the relative shift $\Delta_B = P_b([b_t,b_{t+1}])$ between $b_t$ and $b_{t+1}$. 
However, because of the holes left by the masks, learning the relative shift via a direct matching of the backgrounds would not work. Therefore, we also introduce an inpainting network $\text{N}$. To indicate the masked region to $\text{N}$ we simply fill it with a value out of the image range (we use [-1.1,-1.1,-1.1] as RGB values at the masked pixels). The inpainted regions are then copied to the corresponding backgrounds so that we obtain
$\hat b_j = b_j\odot (1-m_j)+\text{N}(b_j)\odot m_j$, with $j=\{t,t+1\}$. The background reconstructions are then matched with both an $L_1$ norm and a perceptual loss ${\cal L}_\text{VGG}$ based on VGG features \cite{johnson2016perceptual}
\begin{align}
    \textstyle{\cal L}_\text{RECB} = \sum_t \left\| \hat b_{t+1} - \hat b_t \circ \Delta_B \right\|_1 + \lambda_\text{VGG}{\cal L}_\text{VGG}\left(\hat b_{t+1},\hat b_t \circ \Delta_B\right).
\end{align}
Finally, we also have a joint reconstruction loss where we compose the foreground with the estimated foreground shift $\Delta_F$ and the inpainted background with the estimated background shift $\Delta_B$
\begin{align}
    \textstyle{\cal L}_\text{RECJ} = \sum_t \left\|(f_t \odot m_t)\circ \Delta_F + (\hat b_t \circ \Delta_B) \odot (1-m_t \circ \Delta_F) - I_{t+1}\right\|_1.
\end{align}
These losses are all we use to train the mask network $\text{M}$, the inpainting network $\text{N}$ and the pose estimation networks $P_f$ and $P_b$. The inpainting network and the other networks could be further improved, but we find that the choices above are sufficient to obtain accurate segmentation masks and good shift estimates.

\subsection{Local Motion Analysis}

The LMA stage works directly on the foreground frames $f_t$ and $f_{t+1}$. It first shifts $f_t$ with $\Delta_F$. This is done to remove the global shift information from the input frames and to make the action network focus on the local variations. It further encodes the foreground frames with a convolutional neural network $\text{E}$ and obtains $\phi_t = \text{E}(f_t)$, $\tilde\phi_t = \text{E}(f_t\circ \Delta_F)$ and similarly for $\phi_{t+1}= \text{E}(f_{t+1})$. The convolutional feature $\phi_t$ is then projected via $\text{C}$ to give $e_t = \text{C}(\phi_t)$. 

In the action network $\text{A}$ there are a few pre-processing steps. First, both feature maps $\tilde\phi_t$ and $\phi_{t+1}$ are fed to a CNN and flat features $\psi_t$ and $\psi_{t+1}$ are obtained from the resulting feature maps through global average pooling. In CADDY \cite{menapace2021playable}, the actions are determined through a direct difference between Gaussian samples around $\psi_t$ and $\psi_{t+1}$. On average this means that the difference between features of images with the same action must align with the same direction. Although this works very well for CADDY, we find that this may be restrictive, especially if one wants to represent periodic motion (\emph{e.g.}, in our case, an agent walking in place). Thus, we propose to learn a modified mapping of $\psi_{t+1}$ conditioned on $\psi_t$. 
We compute $\psi_{t+1}^i = \text{T}^i(\psi_t, \psi_{t+1}^{i-1})$ with $i=1,\dots,P$, $\text{T}^i$ are bilinear transformations, $\psi_{t+1}^0 = \psi_{t+1}$, and we choose $P=4$.
%We compute $\hat \psi_{t+1} = \text{T}(\psi_t, \cdot) \circ \dots \circ \text{T}(\psi_t,\psi_{t+1})$ through a sequence of four bilinear transforms $\text{T}$ and 
We then compute the action direction $d_t = \psi_{t+1}^P - \psi_t$. Finally, the action $a_t$ 
%$= A(d_t)$ 
is predicted through vector quantization after one additional MLP $\text{U}$ to give $a_t = \text{VQ}[\text{U}(d_t)]$. The vector quantization $\text{VQ}$ relies on $K$ learnable prototype vectors $c_k$, with $k=1,\dots,K$. The method identifies the prototype $c_q$ closest in $L_2$ norm to $\text{U}(d_t)$, \emph{i.e.}, $q = \arg\min_k \|c_k - \text{U}(d_t)\|_2^2$, and uses that as the quantized action $\text{VQ}[\text{U}(d_t)]=c_q$. To train the $\text{VQ}$ prototypes, we use the following loss \cite{van2017neural}
\begin{align}
    {\cal L}_{VQ} = \|\text{sg}[c_q] - U(d_t)\|^2_2 + \lambda_\text{VQ} \| c_q - \text{sg}[U(d_t)]\|^2_2,
\end{align}
where $\lambda_\text{VQ}>0$ is a tuning parameter and $\text{sg}[\cdot]$ denotes stop-gradient.

Now, we have all the inputs needed for the RNN. We introduce an RNN state $s_t$ and feed it together with the encoding $e_t$ as input. Our RNN is split into 6 blocks as in CADDY \cite{menapace2021playable}. Both the global action $\Delta_F$ and the local action $a_t$ are first mapped to embeddings of the same size and then fed to the modulated convolutional layers of the RNN similarly to StyleGAN \cite{karras2021alias}. To differentiate the roles of $\Delta_F$ and $a_t$ we feed the embeddings of $\Delta_F$ to the first two blocks of the RNN and that of $a_t$ to the remaining four blocks. The rationale is that early blocks correlate more with global changes, such as translations, and the later blocks correlate more with local deformations.

Finally, the decoder $\text{D}$ takes the RNN prediction $s_{t+1}$ as input and outputs the frame $\hat f_{t+1} = D_f(s_{t+1})$ and the predicted mask $\hat m_{t+1} = D_m(s_{t+1})$. Moreover, the decoder predicts frames at 3 different scales (as also done in CADDY \cite{menapace2021playable}). We introduce a reconstruction loss for each scale
\begin{align}
   \textstyle {\cal L}_\text{RECU} = \sum_t \left\|\text{sg}[\omega_\text{UNS}] \odot \left(f_{t+1} - \hat f_{t+1}\right)\right\|_1,
\end{align}
where $\forall p\in\Omega$, $\omega_\text{UNS}[p] = \|f_t[p]-f_{t+1}[p]\|_1 + \|\hat f_t[p] - \hat f_{t+1}[p]\|_1$ are weights that enhance the minimization at pixels where the input and predicted foregrounds differ, and also a perceptual loss
\begin{align}
   \textstyle {\cal L}_\text{LMA-VGG} =  {\cal L}_\text{VGG}(f_{t+1},\hat f_{t+1}).
\end{align}
To better learn local deformations, we also introduce a reconstruction loss that focuses on the differences between the foregrounds after aligning them with the estimated relative shifts, \emph{i.e.},
\begin{align}
   \textstyle {\cal L}_\text{RECS} = \sum_t \left\|\text{sg}[\omega_\text{ALIGN}] \odot \left(f_{t+1} - \hat f_{t+1}\right)\right\|_1,
\end{align}
where $\omega_\text{ALIGN}[p] = \|f_t \circ \Delta_F[p] - f_{t+1}[p]\|_1 + \|\hat f_{t+1}[p] - f_{t+1}[p]\|_1$.
To encourage the consistency between the predicted mask $\hat m_{t+1}$ and the mask $m_{t+1}$ obtained from $I_{t+1}$, we also minimize 
\begin{align}
   \textstyle {\cal L}_\text{MSK} = \sum_t \|\hat m_{t+1}-m_{t+1}\|_1.
\end{align}
Moreover, we encourage a cyclic consistency between the encoded features via
\begin{align}
   \textstyle {\cal L}_\text{CYC} = \sum_t \| \text{sg}[\phi_{t+1}] - \text{E}(\hat f_{t+1}) \|_1.
\end{align}

Our final loss consists of a linear combination of all the above losses (both from the GMA and LMA) through corresponding positive scalars $\lambda_\text{VQ}$, $\lambda_\text{LMA-VGG}$, $\lambda_\text{RECU}$, $\lambda_\text{RECS}$, $\lambda_\text{MSK}$, $\lambda_\text{CYC}$, $\lambda_\text{RECF}$, $\lambda_\text{RECB}$, $\lambda_\text{RECJ}$, $\lambda_\text{BIN}$, and $\lambda_\text{SIZE}$.
% \begin{align}
%     % local
%     {\cal L} =& {\cal L}_\text{GLO} + \lambda_\text{LOC} {\cal L}_\text{LOC} + \lambda_\text{VGG} {\cal L}_\text{VGG} + 
%     \lambda_\text{MSK} {\cal L}_\text{MSK}+ \lambda_\text{CYC} {\cal L}_\text{CYC}\\
%     % global
%     &+\lambda_\text{RECF} {\cal L}_\text{RECF} + \lambda_\text{RECB} {\cal L}_\text{RECB} + \lambda_\text{RECJ} {\cal L}_\text{RECJ} + \lambda_\text{BIN} {\cal L}_\text{BIN} + \lambda_\text{SIZE} {\cal L}_\text{SIZE}\nonumber
% \end{align}

\section{Implementation details}

At inference time, GLASS can generate a sequence of frames given only the first one. This setting is slightly different from training, where the model only predicts the next frame given the previous one. In order to prepare the model for test time, we adopt the mixed training procedure (Teacher Forcing) also used in \cite{menapace2021playable}. That is, we select a video duration $T_f$, $0 < T_f < T$, and if $t \le T_f$ we feed the encodings of the real frames to the RNN, otherwise if $t > T_f$ we use the encodings of the reconstructed frames. During the training we gradually decrease $T_f$ to $1$ and increase $T$ to adapt the network to the generation of longer sequences.
To speed up the convergence, we pretrain the GMA component for 3000 iterations.
%\textcolor{red}{To ensure meaningful input to the LMA module and hence, better convergence, we pretrain the GMA component for the first 3000 iterations.}
The coefficients before the loss terms are estimated on the training set. We found that the selected configuration works well across all datasets.  The models are trained using the Adam optimizer \cite{kingma2014adam} with a learning rate equal to $0.0004$ and weight decay $10^{-6}$. For more details, see the supplementary material.

\section{W-Sprites dataset}\label{sec:wsprites}

In order to assess and ablate the components of GLASS,
we build a synthetic video dataset of cartoon characters acting on a moving background.
We call the dataset W-Sprites (for Walking Sprites). Each sequence is generated via the following procedure.
First, one of 1296 different characters is sampled from the Sprites dataset \cite{liberatedpixelcup,li2018disentangle,universallpcspritesheet}.
This character is then animated in two stages. A random walk module produces a sequence of global coordinates of the sprite within 
a $96\times128$ resolution frame. We then sample one of 9 local actions conditioned on the shift induced by the global motion component. 
Those actions include: \texttt{walk front}, \texttt{walk left}, \texttt{walk right}, \texttt{spellcast front}, \texttt{spellcast left}, \texttt{spellcast right}, \texttt{slash front}, \texttt{slash left}, and  
\texttt{slash right}. The intuition under conditioning is that the global actions and the local ones should be correlated for more realism. 
For instance, when the global action module dictates the right shift, the only possible local action should be \texttt{walk right}.
Analogously, the left shift induces the \texttt{walk left} action. The up and down shifts are animated with the \texttt{walk front} action.
The remaining actions are used to animate the static sprite.
To incorporate more generality and to reduce the gap with real data, we apply an independent random walk to the background image (this simulates camera motion). We use a single background image sampled from the ``valleys'' class of ImageNet \cite{deng2009imagenet}.
Each video in the W-Sprites dataset is annotated with per frame actions (\emph{i.e.}, global shifts and local action identifiers), background shifts and character masks.
We show sequence samples from our dataset in Fig.~\ref{fig:wsprites} (to play the videos view the pdf with Acrobat Reader). The dataset contains 10 to 90 frames long videos per sprite.
For testing purposes we split the dataset into training and  validation sets. The validation set contains sprites never seen during training. The training set is approximately 8 times larger than the validation one. For more details, see the supplementary material.

% \begin{figure}[t] 
%     \centering
%     \includegraphics[width=\textwidth]{Figures/wsprites.png}
%     \caption{Sample sequences from our W-Sprites dataset.\label{fig:wsprites}}
% \end{figure}
\begin{table}[t]
\centering
\caption{Global Motion Analysis (GMA). mIoU evaluations}
\label{tab:global-ablationsmIoU}
\begin{tabular*}{\textwidth}{@{}l@{\extracolsep{\fill}}ccccccc@{}}
\toprule
{Configuration} &  
\cancel{${\cal L}_\text{RECB}$} &
\cancel{${\cal L}_\text{RECF}$} &
\cancel{${\cal L}_\text{SIZE}$} &
\cancel{${\cal L}_\text{RECJ}$} &
\cancel{${\cal L}_\text{BIN}$}  &
\cancel{${\cal L}_\text{VGG}$}  &
GLASS\\
\midrule
{mIoU} & 0.01 & 0.08 & 0.08 & 0.08 & 0.87 & \textbf{0.89} & 0.88\\
\bottomrule
\end{tabular*}
\end{table}

\section{Ablations}

In this section we separately ablate the global and local components of GLASS. We run the ablations on {\bf W-Sprites}, which has been introduced in section~\ref{sec:wsprites}. \\
\noindent\textbf{GMA ablations.} For the global motion analysis, we assess the impact of each loss 
function. Different loss terms are sequentially switched off and the performance of the model trained without those terms is reported. Given that W-Sprites is fully annotated, we propose several metrics to evaluate the training. First, we calculate the mean intersection over union ($\text{mIoU}$)  between the ground truth and the predicted segmentation masks. Table~\ref{tab:global-ablationsmIoU} shows that the VGG loss seems to slightly hurt the segmentation performance. However, as shown in Table~\ref{tab:global-ablationsShifts} the VGG loss benefits the shift estimation.
Notice that in Table~\ref{tab:global-ablationsShifts} we report only the cases where the masks are good enough ($\text{mIoU} > 0.8$). For the shift errors we show the $L_2$ norm of the difference between the ground truth foreground/background shift and the predicted one (in pixels). We also show the accuracy of the predicted foreground/background shift directions ($\measuredangle$-ACC). The direction is considered to be correctly predicted if the angle between the ground truth and the predicted shifts is less than 45\degree. %Tables~\ref{tab:global-ablationsmIoU} and \ref{tab:global-ablationsShifts} provide the evaluation results on the considered global motion component ablations. 
Each model is trained for 60K iterations with a batch size of 4. The results are calculated on the validation set.

\begin{table}[t]
\centering
\caption{Global Motion Analysis (GMA). Shift error estimation}
\label{tab:global-ablationsShifts}
\begin{tabular*}{\textwidth}{@{}l@{\extracolsep{\fill}}|cccc|cccc@{}}
\toprule
\multicolumn{1}{c|}{\multirow{2}{*}{{Configuration}}} & %\multirow{2}{*}{\textbf{mIoU}} &
\multicolumn{4}{c|}{{Background Shift Error}}                          & \multicolumn{4}{c}{{Foreground Shift Error}}                              \\
\multicolumn{1}{c|}{}  % &  
& {mean} & {min}  & {max}  & {$\measuredangle$-ACC} & {mean} & {min}  & {max}  & {$\measuredangle$-ACC} \\ \hline
% w/o ${\cal L}_\text{RECB}$                                                    & 0.01                           & -             & -             & -             & -                             & -             & -             & -             & -                             \\
% w/o ${\cal L}_\text{RECF}$                                                    & 0.08                           & -             & -             & -             & -                             & -             & -             & -             & -                             \\
% w/o ${\cal L}_\text{SIZE}$                                             & 0.08                           & -             & -             & -             & -                             & -             & -             & -             & -                             \\
% w/o ${\cal L}_\text{RECJ}$                                                   & 0.08                           & -             & -             & -             & -                             & -             & -             & -             & -                             \\
\cancel{${\cal L}_\text{BIN}$}                                     %      & 0.87                           
& 0.55          & 0.01          & 1.16          & \textbf{1.00}                          & 4.46          & 0.05          & 8.50          & \textbf{1.00}                          \\
\cancel{${\cal L}_\text{VGG}$}                                     %          & \textbf{0.89}                  
& 0.52          & 0.02          & 0.90          & \textbf{1.00}                 & 4.38          & \textbf{0.01} & 8.51          & \textbf{1.00}                 \\
GLASS                                                         %& 0.88  
& \textbf{0.51} & \textbf{0.00} & \textbf{0.86} & \textbf{1.00}                 & \textbf{4.34} & 0.02          & \textbf{8.32} & \textbf{1.00}                          \\ \bottomrule
\end{tabular*}
\end{table}
\begin{table}[t]
    \centering
    \caption{Local Motion Analysis (LMA). Component ablation results}
    \label{tab:local-ablations}
    \begin{tabular*}{\textwidth}{@{}l@{\extracolsep{\fill}}cccccccc@{}}
    \toprule%\noalign{\smallskip}
        Configuration & LPIPS$\downarrow$ & FID$\downarrow$ & FVD$\downarrow$ & $\text{mIoU}_\text{RE}\uparrow$ & $\text{NMI}_\text{G}\uparrow$ & $\text{NMI}_\text{S}\uparrow$ & $\text{CON} \downarrow$\\
    %\noalign{\smallskip}
    \hline%\noalign{\smallskip}
        Plain directions & \textbf{0.417} & \textbf{26.0} & 192 & 0.83 & 0.14 & 0.17 & 0.03\\
        %Plain directions & 0.112 & 40.2 & 268 & 0.83 & 0.14 & 0.17 & 0.03\\
        Gumbel & \textbf{0.417} & 30.4 & 304 & 0.84 & 0.00 & 0.02 & 0.02\\
        %Gumbel & 0.123 & 48.9 & 418 & 0.84 & 0.00 & 0.02 & \underline{0.02}\\
        No modulated convs & \underline{0.420} & 26.8 & \underline{173} & \textbf{0.89} & \underline{0.35} & \underline{0.38} & 0.03\\
        %No modulated convs & 0.121 & 42.2 & 236 & \textbf{0.89} & \underline{0.35} & \underline{0.38} & 0.03\\
        Joint input & 0.421 & \underline{26.3} & \textbf{156} & \textbf{0.89} & 0.34 & 0.37 & 0.03\\
        %Joint input & 0.120 & 41.2 & 228 & \textbf{0.89} & 0.34 & 0.37 & 0.03\\
        \cancel{${\cal L}_\text{RECS}$} & 0.422 & 29.3 & 212 & \textbf{0.89} & 0.29 & 0.30 & \textbf{0.01} \\
        %\cancel{${\cal L}_\text{RECS}$} & 0.125 & 43.2 & 292 & \textbf{0.89} & 0.29 & 0.30 & \textbf{0.01} \\
    %\noalign{\smallskip}
    \hline%\noalign{\smallskip}
        GLASS 200K & 0.421 & 27.8 & 188 & \underline{0.88} & {\bf 0.39} & {\bf 0.41} & \underline{0.02} \\
        \hline
        %GLASS 800K & - & - & - & \underline{0.88} & {\bf 0.39} & {\bf 0.41} & \underline{0.02} \\
        GLASS 470K & 0.417 & 21.4 & 137 & 0.93 & 0.39 & 0.40 & 0.01 \\
        %GLASS 1.8m & 0.104 & 30.9 & 176 & 0.93 & 0.39 & 0.40 & 0.01 \\
    %\noalign{\smallskip}\hline
    \bottomrule
    \end{tabular*}
\end{table}

\noindent\textbf{LMA ablations.} For the local motion analysis module we specifically design 5 cases that differ from GLASS in its essential components and show the results in Table~\ref{tab:local-ablations}. First, we evaluate swapping the modified mapping $T$ of the features $\psi_{t+1}$ for the direct difference between the features $\psi_{t+1}-\psi_t$ (as done in CADDY \cite{menapace2021playable}). We refer to this configuration as ``Plain directions''. Second, we replace the vector quantization with an MLP that predicts the distribution over actions followed by the Gumbel-Softmax trick to sample a discrete action identifier. We name this model ``Gumbel''. We also ablate the impact of using modulated convolutional layers by feeding the action embeddings as normal inputs to common convolutional blocks. This cases is referred to as ``No modulated convs''. Also we consider the case where we feed the global and local action embeddings jointly to all RNN blocks instead of separate ones. We refer to this case as ``Joint input''. The last case that we evaluate for the ablations is the model trained without ${\cal L}_\text{RECS}$. All the models are trained for 200K iterations with a batch size of 4. Additionally we report the metrics of GLASS trained for 470K iterations.

\begin{figure}[t]
\centering
\begin{minipage}[b]{.45\textwidth}
  \centering
  \includegraphics[width=0.95\linewidth]{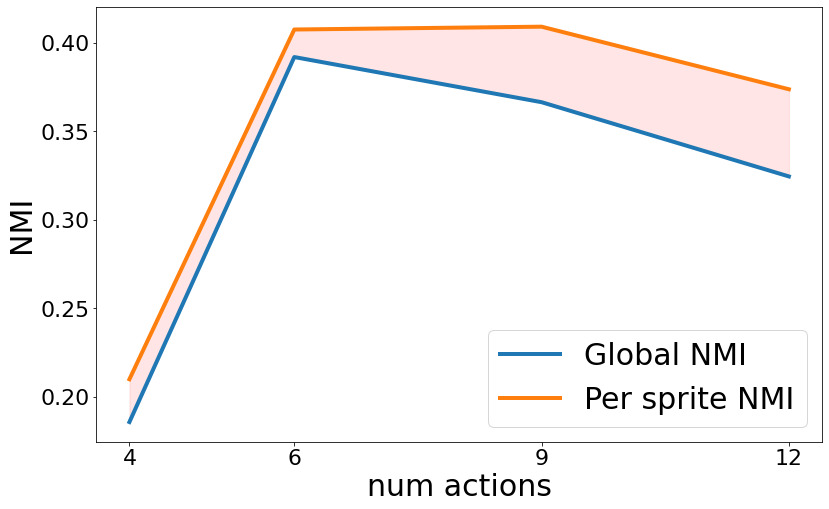}\captionof{figure}{Ablation of the number of actions fitted during the training of GLASS on the \emph{W-Sprites} dataset.}
  \label{fig:num_actions}
\end{minipage}\hspace{3em}
\begin{minipage}[b]{.45\textwidth}
  \centering
      \begin{tabular}{c}
        Tennis source video \\
        \animategraphics[width=0.95\textwidth]{7}{Figures/tennis_real/real_}{0}{15}\\
        Tennis target video \\
        \animategraphics[width=0.95\textwidth]{7}{Figures/tennis_transfer/transfer_}{0}{14}
    \end{tabular}
  \captionof{figure}{Example of transferring an action sequence decoded from the source video to the target. To play view the paper in Acrobat Reader.}
  \label{fig:transfer}
\end{minipage}
\end{figure}

Following CADDY \cite{menapace2021playable}, we generate the sequences from the first frames of the validation videos conditioned on the actions inferred from the remaining frames. We measure FID \cite{heusel2017gans}, FVD \cite{unterthiner2018towards} and LPIPS \cite{zhang2018unreasonable} scores on the generated sequences to asses the quality of the generated videos. Additionally we segment the reconstructed sequences and report the mean IoU with the ground truth masks to asses the ability of the RNN to condition on the input global and local action embeddings. We also propose to use the normalized mutual information score (NMI) between the ground truth and inferred local actions
\begin{align}
\textstyle    \text{NMI}(X, Y) = \frac{2 I(X, Y)}{H(X) + H(Y)},
\end{align}
where $I(X, Y)$ is the mutual information between $X$ and $Y$ and $H(X)$ is the entropy of $X$.
One appealing advantage of NMI for GLASS is that NMI is invariant to permutations of the labels. %That means that one does not have to explicitly find the permutation that yields the best accuracy. 
Another advantage of using NMI is that NMI does not require the distributions to have the same number of actions. Thus, even with a given known number of ground truth actions, the model can be trained and assessed with a different number of actions. 
Indeed, the decomposition of a sequence into actions is not unique.
%Indeed, a specific sequence can be generated with a variety of different sets of actions. 
For instance the \texttt{walk left} action can be decomposed into \texttt{turn left} and \texttt{walk}.
We introduce two different protocols of measuring NMI. First, we classify all the pairs of successive frames to different actions. Then the global $\text{NMI}_\text{G}$ is computed between the ground truth actions and those predictions. Additionally, we average the per sprite NMI scores to obtain $\text{NMI}_\text{S}$. Normally $\text{NMI}_\text{S} > \text{NMI}_\text{G}$. However, if the gap is large enough, this indicates the overfitting and the lack of consistency of the learned actions across different sprites. Therefore, we also report the consistency metric $\text{CON} = \text{NMI}_\text{S} - \text{NMI}_\text{G}$. As a reference we use the $\text{NMI}_\text{RAND}$, that is the NMI measured between the ground truth actions and random actions. The results are provided in  Table~\ref{tab:local-ablations}. Given that $\text{NMI}_\text{RAND} = 0.02$ on the W-Sprites test set, the full GLASS configuration with an NMI of $0.41$ shows that the model estimates action sequences with a high correlation to the ground truth actions. 
%We summarize the results of this ablation in Table~\ref{tab:local-ablations}.
Furthermore, we ablate the number of actions $K$ used to train GLASS. In Fig.~\ref{fig:num_actions} one can see that $K=6$ is optimal in both NMI and CON.

% \begin{figure}[t]
%     \centering
%     \includegraphics[width=0.6\textwidth]{Figures/num_actions.png}
%     \caption{Ablation of the number of actions fitted during the training of GLASS.}
%     \label{fig:num_actions}
% \end{figure}

% \begin{figure}[t]
%     \centering
%     \begin{tabular}{cc}
%         Source video & Target video \\
%         \animategraphics[width=0.35\textwidth]{7}{Figures/tennis_real/real_}{0}{15} &
%         \animategraphics[width=0.35\textwidth]{7}{Figures/tennis_transfer/transfer_}{0}{14}
%     \end{tabular}
%     \caption{Example of transferring an action sequence decoded from the source video to the target. To play use acrobat reader.}
%     \label{fig:transfer}
% \end{figure}

\begin{table}[t]
    \centering
    \caption{\emph{BAIR} dataset evaluation}
    \label{tab:bair}
    \begin{tabular*}{\textwidth}{l@{\extracolsep{\fill}}ccc@{}}
    \toprule%\noalign{\smallskip}
        Method & LPIPS$\downarrow$ & FID$\downarrow$ & FVD$\downarrow$\\
    %\noalign{\smallskip}
    \hline
        MoCoGAN \cite{tulyakov2018mocogan} & 0.466 & 198 & 1380 \\
        MoCoGAN+ \cite{menapace2021playable} & 0.201 & 66.1 & 849 \\
        SAVP \cite{lee2018stochastic} & 0.433 & 220 & 1720 \\
        SAVP+ \cite{menapace2021playable} & \underline{0.154} & \underline{27.2} & \underline{303} \\
        \hline
        Huang et al. \cite{huang2021layered} w/ \emph{non-param} control & 0.176 & 29.3 & 293 \\
    %\noalign{\smallskip}
    \hline%\noalign{\smallskip}
        CADDY \cite{menapace2021playable} & 0.202 & 35.9 & 423 \\
        {Huang et al. \cite{huang2021layered} w/ \emph{positional} control}  & 0.202 & 28.5 & 333 \\
        Huang et al. \cite{huang2021layered} w/ \emph{affine} control & 0.201 & 30.1 & \textbf{292} \\
            %\noalign{\smallskip}
    \hline%\noalign{\smallskip}
        GLASS & \textbf{0.118} & \textbf{18.7} & 411 \\
        %GLASS & \textbf{0.120} & \textbf{18.8} & 420 \\
    %\noalign{\smallskip}
    \bottomrule
    \end{tabular*}
\end{table}
\begin{table}[t]
    \centering
    \caption{\emph{Tennis} dataset evaluation}
    \label{tab:tennis}
    \begin{tabular*}{\textwidth}{@{}l@{\extracolsep{\fill}}ccccc@{}}
    \toprule%\noalign{\smallskip}
        Method & LPIPS$\downarrow$ & FID$\downarrow$ & FVD$\downarrow$ & ADD$\downarrow$ & MDR$\downarrow$\\
    %\noalign{\smallskip}
    \hline
        MoCoGAN \cite{tulyakov2018mocogan} & 0.266 & 132 & 3400 & 28.5 & 20.2 \\
        MoCoGAN+ \cite{menapace2021playable} & 0.166 & 56.8 & 1410 & 48.2 & 27.0 \\
        SAVP \cite{lee2018stochastic} & 0.245 & 156 & 3270 & 10.7 & 19.7 \\
        SAVP+ \cite{menapace2021playable} & 0.104 & 25.2 & 223 & 13.4 & 19.2 \\
        \hline        
        Huang et al. \cite{huang2021layered} w/ \emph{non-param} control & 0.100 & 8.68 & 204 & 1.76 & 0.306 \\
    %\noalign{\smallskip}
    \hline%\noalign{\smallskip}
        CADDY \cite{menapace2021playable} & \underline{0.102} & 13.7 & 239 & 8.85 & 1.01 \\
        {Huang et al. \cite{huang2021layered} w/ \emph{positional} control}  & 0.122 & \underline{10.1} & \underline{215} & 4.30 & \underline{0.300} \\
        Huang et al. \cite{huang2021layered} w/ \emph{affine} control & 0.115 & 11.2 & \textbf{207} & \underline{3.40} & 0.317 \\
    %\noalign{\smallskip}
    \hline%\noalign{\smallskip}
    % nearest acc
    GLASS & \textbf{0.046} & \textbf{7.37} & 257 & \textbf{2.00} & \textbf{0.214} \\
    % accumulated shifts
    % GLASS & \textbf{0.051} & \underline{10.3} & 305 & \textbf{2.00} & \textbf{0.275} \\
    % sequential shifts
    %GLASS & \textbf{0.067} & 17.2 & 405 & \textbf{2.00} & \underline{0.306} \\
    %\noalign{\smallskip}
    \bottomrule
    \end{tabular*}
\end{table}

\section{Experiments}

\begin{figure}[t]
    \centering
    \begin{tabular}{@{}c@{\hspace{2mm}}c@{\hspace{2mm}}c@{\hspace{2mm}}c@{\hspace{2mm}}c@{\hspace{2mm}}c@{\hspace{2mm}}c@{\hspace{2mm}}c@{}}
        \includegraphics[width=0.4cm]{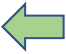} &
        \includegraphics[width=0.4cm]{Figures/actions/left.png} &
        \includegraphics[width=0.4cm]{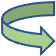} &
        \includegraphics[width=0.4cm]{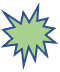} &
        \includegraphics[width=0.4cm]{Figures/actions/left.png} &
        \includegraphics[width=0.3cm]{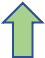} &
        \includegraphics[width=0.4cm]{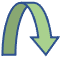} &
        \includegraphics[width=0.4cm]{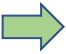} \\
        \animategraphics[width=0.11\textwidth]{7}{Figures/walk/walk_}{0}{27} & 
        \includegraphics[width=0.11\textwidth]{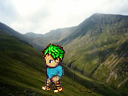} &
        \includegraphics[width=0.11\textwidth]{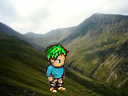} & 
        \includegraphics[width=0.11\textwidth]{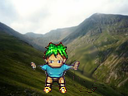} & 
        \includegraphics[width=0.11\textwidth]{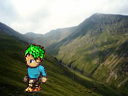} & 
        \includegraphics[width=0.11\textwidth]{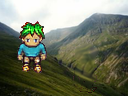} &
        \includegraphics[width=0.11\textwidth]{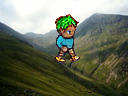} & 
        \includegraphics[width=0.11\textwidth]{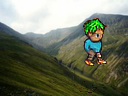}\\
        \texttt{walk} & \texttt{walk} & \texttt{turn} & \texttt{spell-} & \texttt{walk} & \texttt{walk} & \emph{jump} & \texttt{walk}\\
        \texttt{left} & \texttt{left} & \texttt{right} & \texttt{cast} & \texttt{left} & \texttt{up} &  & \texttt{right}
    \end{tabular}
    \caption{A sequence generated with GLASS trained on the \emph{W-Sprites} dataset. Note that the level of control provided by GLASS allows to generate unseen motion such as \emph{jump}. Use Acrobat Reader to play the first frame.}
    \label{fig:control}
\end{figure}

We evaluate GLASS on 3 datasets. For synthetic data we use \textbf{W-Sprites}. For real data we use: 1) the {\bf Tennis Dataset} and 2) the {\bf BAIR Robot Pushing Dataset}. The Tennis Dataset was introduced in \cite{menapace2021playable} and contains around $900$ videos extracted from $2$ Tennis matches from YouTube at $96\times256$ pixel resolution. The videos are cropped to contain only one half of the court, so that only one player is visible. The BAIR Robot Pushing Dataset \cite{ebert2017self} contains around 44K clips of a robot arm pushing toys on a flat square table at $256\times256$ pixel resolution.\\
\noindent\textbf{Baselines.} 
We compare to CADDY \cite{menapace2021playable}, since it allows frame-level playable control, and  to Huang et al.~\cite{huang2021layered}. However, we do not explicitly compare to the non-parametric control model \cite{huang2021layered}, since it requires a prior knowledge of the future agent masks and also it lacks the ability to control the agent through discrete actions (playability).
%Our main competitor is CADDY \cite{menapace2021playable}, since it allows frame-level playable control. We also consider Huang et al. \cite{huang2021layered}, however we do not explicitly compare to their non-parametric control model, since it lacks the playability and requires a prior knowledge of the future agent masks. 
We also report the metrics on other conditional video generation models such as MoCoGAN \cite{tulyakov2018mocogan}, SAVP \cite{lee2018stochastic} and their large scale versions from \cite{menapace2021playable}.\\
\noindent\textbf{Quantitative analysis.}
Following \cite{menapace2021playable} we evaluate GLASS on the video reconstruction task. Given a test video, we use GMA to predict the global shifts and LMA to estimate the discrete actions performed along the video. Further, the agent is segmented using the masking network and the foreground is animated and pasted back to the shifted background using both global and local actions to reconstruct the whole sequence from the first frame.
We report FID, FVD and LPIPS scores on the generated videos. On the Tennis dataset we additionally report the Average Detection Distance (ADD) and the Missing Detection Rate (MDR) suggested in \cite{menapace2021playable}. Those metrics are supposed to assess the action space quality by detecting the tennis player with a pretrained human detector and by comparing the locations of the detected agents in the ground truth and generated sequences. On BAIR (see Table~\ref{tab:bair}) our model performs almost 40\% better in terms of frame-level quality, but lacks in FVD compared to \cite{huang2021layered}. However, it is still slightly better than CADDY. On the Tennis dataset (see Table~\ref{tab:tennis}) GLASS is around 50\% better than the closest competitor in LPIPS, almost 30\% better in FID, but loses in FVD. However, GLASS provides finer control over the agent according to ADD and MDR.\\
\noindent\textbf{Qualitative analysis.}
A trained GLASS allows a detailed control of the agent. On W-Sprites, we find that the LMA discovers such actions as \texttt{turn right}, \texttt{turn left}, \texttt{turn front}, \texttt{spellcast} and \texttt{slash}. Note that despite the difference between the discovered set of actions and the ground truth, all videos in the training set can be generated with this reduced set of actions 
% this part seems unfinished
(see Fig.~\ref{fig:control}). On Tennis we found that the local actions mostly correspond to some leg movements. On BAIR the LMA component discovers some small local deformations such as the state of the manipulator (closed or open).

\begin{figure}[t]
    \centering
    \begin{tabular}{@{}ccccc@{}}
    %\centering
    \includegraphics[width=0.088\textwidth]{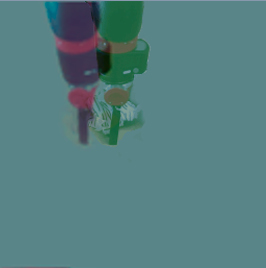}\includegraphics[width=0.088\textwidth]{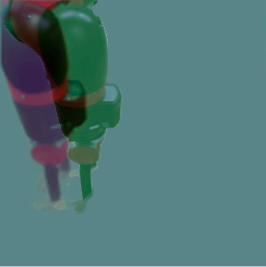}&
    \includegraphics[width=0.088\textwidth]{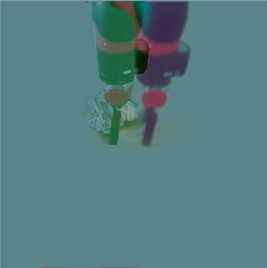}\includegraphics[width=0.088\textwidth]{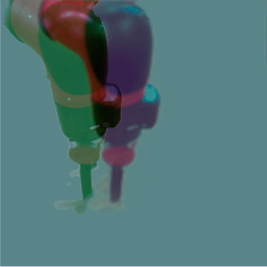}&
    \includegraphics[width=0.088\textwidth]{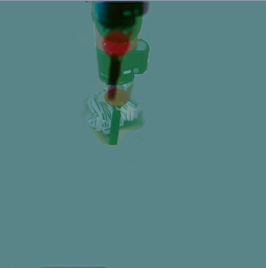}\includegraphics[width=0.088\textwidth]{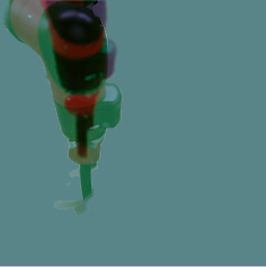}&
    \includegraphics[width=0.088\textwidth]{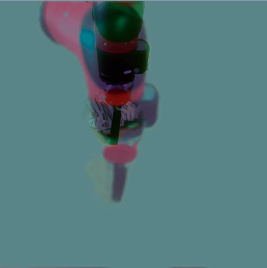}\includegraphics[width=0.088\textwidth]{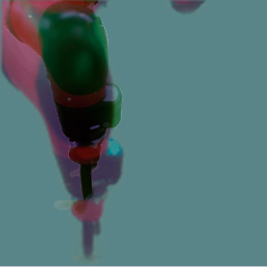}&
    \includegraphics[width=0.088\textwidth]{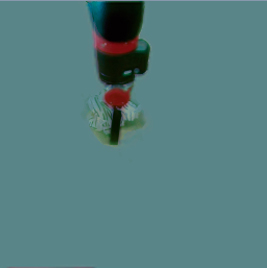}\includegraphics[width=0.088\textwidth]{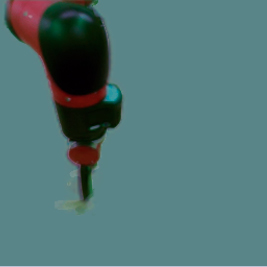}\\
    \includegraphics[width=0.176\textwidth]{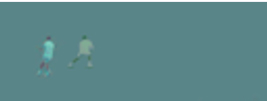}&
    \includegraphics[width=0.176\textwidth]{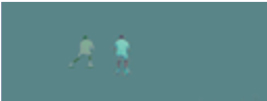}&
    \includegraphics[width=0.176\textwidth]{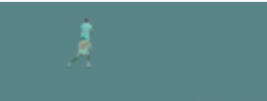}&
    \includegraphics[width=0.176\textwidth]{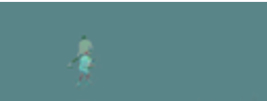}&
    \includegraphics[width=0.176\textwidth]{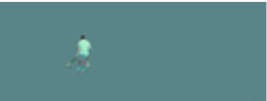}\vspace{-2.5mm}
    \\
    \includegraphics[width=0.176\textwidth]{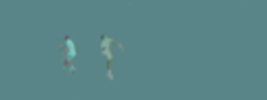}&
    \includegraphics[width=0.176\textwidth]{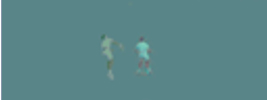}&
    \includegraphics[width=0.176\textwidth]{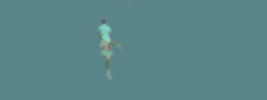}&
    \includegraphics[width=0.176\textwidth]{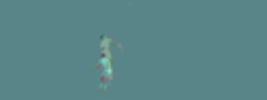}&
    \includegraphics[width=0.176\textwidth]{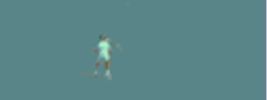}\\
    left&
    right&
    up&
    down&
    no motion
    \end{tabular}
    \caption{Learned global actions on the \emph{BAIR} and \emph{Tennis} datasets (foreground-only generation). We use the green channel for the given initial foreground and the red channel for the foreground generated with the selected action. For both datasets we show 2 examples to demonstrate the action consistency. 
    %\textcolor{red}{Overlayed are shown the 1st and the 5th images in sequence. The initial frame is transparent. Each column applies the same global action, from left to right: \texttt{right}, \texttt{left}, \texttt{down}, \texttt{up}, \texttt{no motion}. Consistency of actions is shown on two sample images per action and dataset.}
    }
    \label{fig:matrix}
\end{figure}

In Fig.~\ref{fig:matrix}, we provide visual examples of the GLASS global action space. Given two different starting foregrounds from the BAIR and Tennis datasets (shown in the green channel), we show the generated foregrounds (in the red channel) after applying the \texttt{right}, \texttt{left}, \texttt{down}, \texttt{up} and \texttt{no motion} global shifts.
We can also see that global actions apply consistently across different initial foregrounds. To show that the learned action space is consistent across different agents also in their fine-grained dynamics we use GLASS to transfer (both global and local) motion from one video to another. We first extract the sequence of actions in the first video using the GMA and LMA components of GLASS and then sequentially apply these actions to the first frame of the second video. In Fig.~\ref{fig:transfer}, we demonstrate it on the Tennis dataset.

\begin{figure}[H]
    \centering
    \begin{tabu}{c@{\hspace{1mm}}c}
        \includegraphics[width=.10\textwidth]{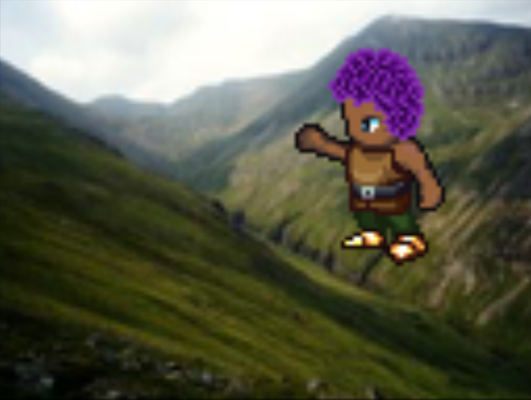}&
        \multirow{2}[2]{*}[3.2mm]{\includegraphics[width=.12\textwidth]{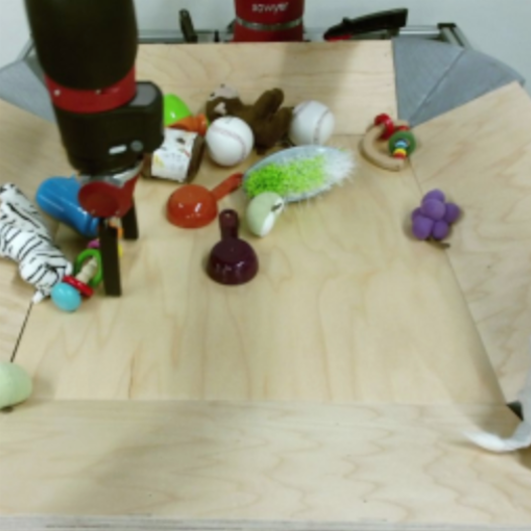}}\\
        \includegraphics[width=.10\textwidth]{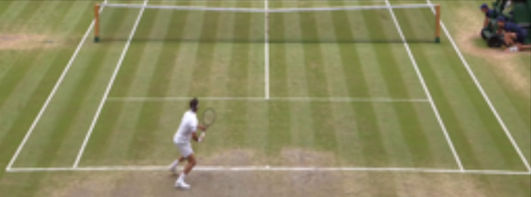}&\\
        \multicolumn{2}{c}{original}
    \end{tabu}\hspace{-1.4mm}\begin{tabu}{c@{\hspace{1mm}}c}
        \includegraphics[width=.10\textwidth]{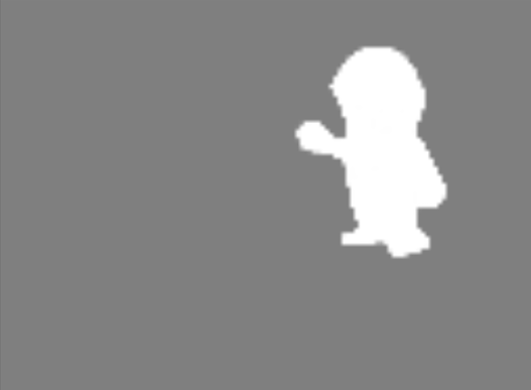}&
        \multirow{2}[2]{*}[3.0mm]{\includegraphics[width=.12\textwidth]{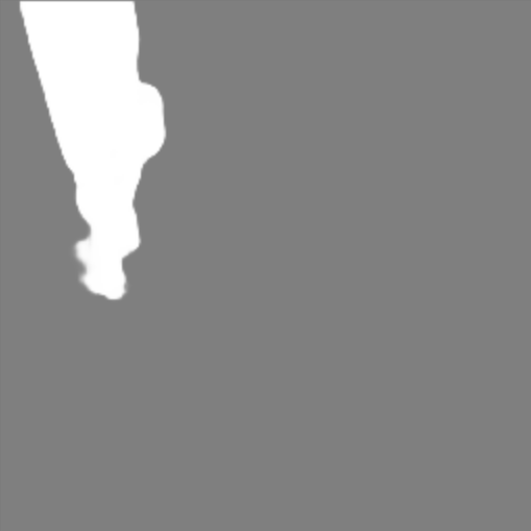}}\\
        \includegraphics[width=.10\textwidth]{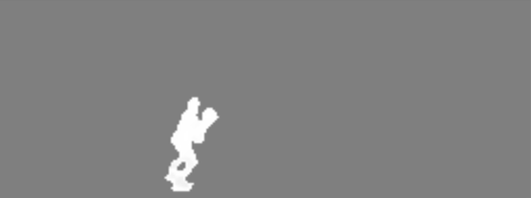}&\\
        \multicolumn{2}{c}{mask}
    \end{tabu}\hspace{-1.4mm}\begin{tabu}{c@{\hspace{1mm}}c}
        \includegraphics[width=.10\textwidth]{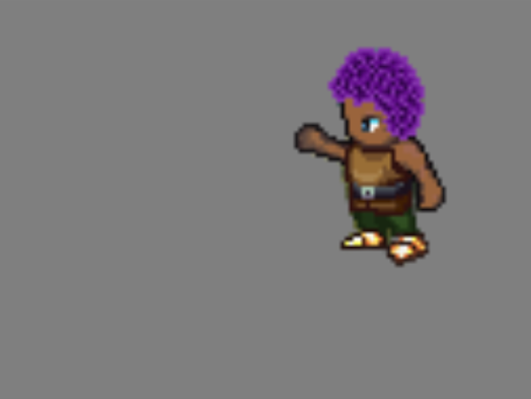}&
        \multirow{2}[2]{*}[3.15mm]{\includegraphics[width=.12\textwidth]{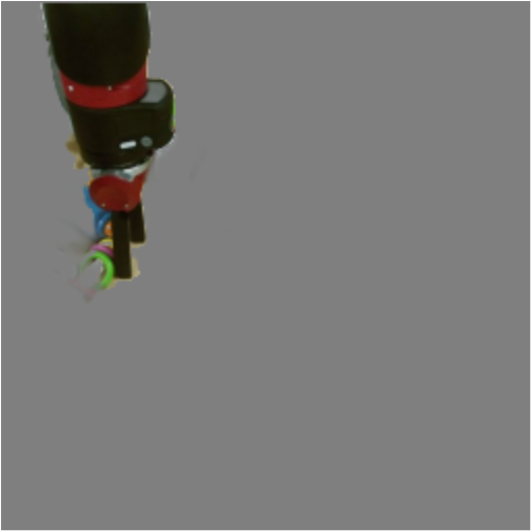}}\\
        \includegraphics[width=.10\textwidth]{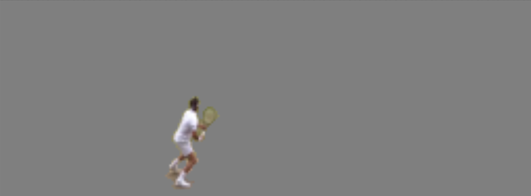}&\\
        \multicolumn{2}{c}{foreground}
    \end{tabu}\hspace{-1.4mm}\begin{tabu}{c@{\hspace{1mm}}c}
        \includegraphics[width=.10\textwidth]{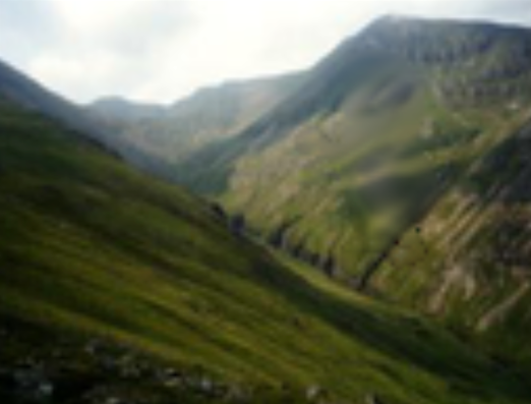}&
        \multirow{2}[2]{*}[3.2mm]{\includegraphics[width=.12\textwidth]{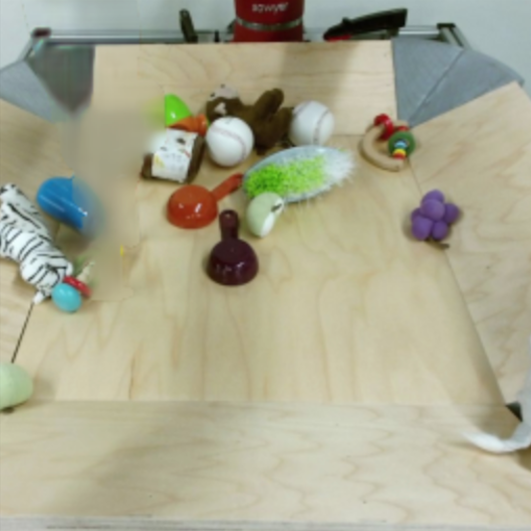}}\\
        \includegraphics[width=.10\textwidth]{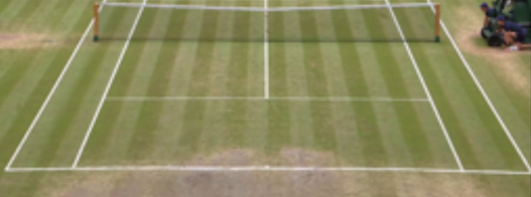}&\\
        \multicolumn{2}{c}{background}
    \end{tabu}     
    \caption{Sample outputs from our GMA module. From left to right: original image, predicted segmentation, foreground, and inpainted background.}
    \label{fig:layers}
\end{figure}

Finally, in Fig.~\ref{fig:layers} we provide some sample outputs from our GMA module on test images from all three datasets. Given an input image, we can see that the segmentation network learns to extract accurate masks with which one can obtain high quality foreground images. These are necessary to model local dynamics. The inpainting of the background is sufficiently accurate to separate the two layers.  %We demonstate the precision of the segmentation and inpainting of our model. %However, as mentioned above, the inpainter can be further improved separately as a postprocessing step.
For more visual examples, please see the supplementary material.

% \section{Discussion and Future work}

% Some words about limitations. 

\section{Conclusions}

GLASS is a novel generative model with a global and local action space that enables a fine-grained modeling and control of dynamics not available in prior work. GLASS is trained in a completely unsupervised manner. We also introduce W-Sprites, a novel dataset for the evaluation of action identification and generation. Our experimental evaluation shows that GLASS learns consistent, and thus transferrable, action representations and is able to synthesize realistic videos with arbitrary action policies.

%iii) We demonstrate GLASS on both synthetic and real datasets and show that it can: 1) segment an agent from its environment and estimate its global shift over time; 2) learn a disentangled action space that is consistent across agents; 3) decode videos into sequences of actions; 4) synthesize realistic videos under the guidance of a novel action policy.

\noindent\textbf{Acknowledgements}

This work was supported by grant 188690 of the Swiss National Science Foundation.

\clearpage

\bibliographystyle{splncs04}
\bibliography{references}

%%%%%%%%%%%%%%%%%%%%%%%%%%%%%%%%%%%%%%%%%%%%%%%%%%%%%%%%%%%%

%%%%%%%%%%%%%%%%%%%%%%%%%%%%%%%%%%%%%%%%%%%%%%%%%%%%%%%%%%%%

\clearpage

\appendix

\section{Appendix}

In the main paper we present GLASS, a method for Global and Local Action-driven Sequence Synthesis. GLASS, trained on unlabeled video sequences, allows to animate an input image at test time. The method builds a global and local action representation that is used to generate transitions of the segmented foreground sequences. Moreover, we introduced a novel dataset (W-Sprites) with a predefined action space for analysis. This supplementary material provides details and visual examples that could not be included in the main paper due to the space limitations. In section~\ref{sec:impl} we describe the implementation details, such as network architecture and training parameters. Section~\ref{sec:data} provides details on the dataset generation protocol. In Section~\ref{sec:vis} we include more visual examples of the evaluation of our method. Further details, the code and example videos will be made available on github.

\section{Implementation}\label{sec:impl}

In this section we report further details regarding the implementation of GLASS. 

\subsection{Network architecture}

In our code we mostly adopt convolutional blocks from the publicly available implementation of CADDY~\cite{menapace2021playable}. Those include residual blocks, upsampling and downsampling blocks. The blocks mainly incorporate Leaky-ReLu activations and Batch Normalization layers. Exceptions are the blocks that output masks (sigmoid activation), the blocks that output images ($\tanh$ activation) and the LSTM blocks (sigmoid and $\tanh$ activations) \cite{hochreiter1997long}.

\begin{figure}[t]
    \centering
    \begin{tabular}{@{}c|c|c@{}}
        Masking Network & Shift Predictor & Inpainter \\
        \includegraphics[height=4cm]{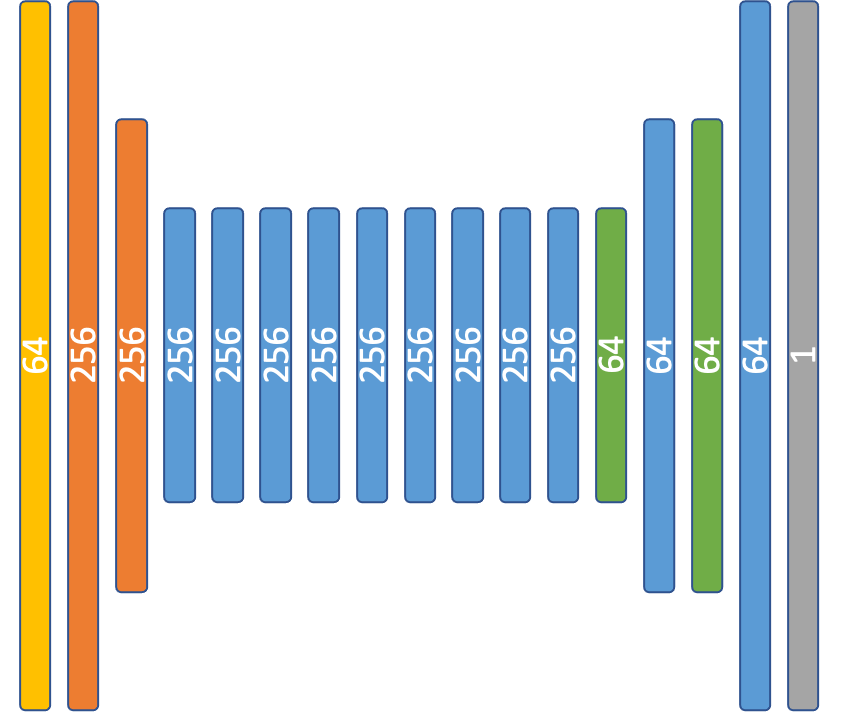} &
        \includegraphics[height=4cm]{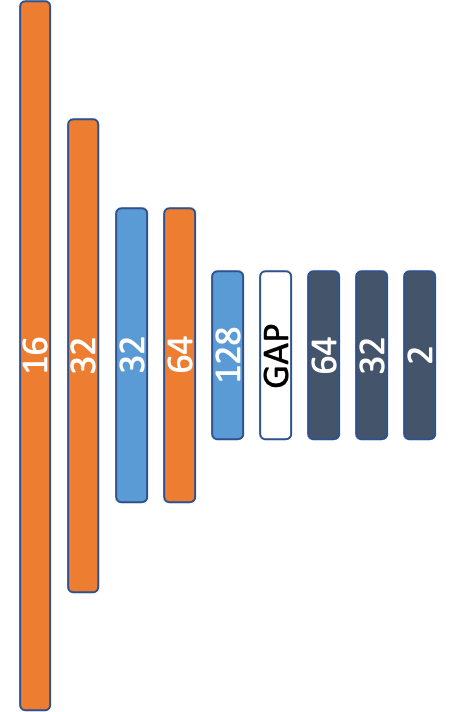} & 
        \includegraphics[height=4cm]{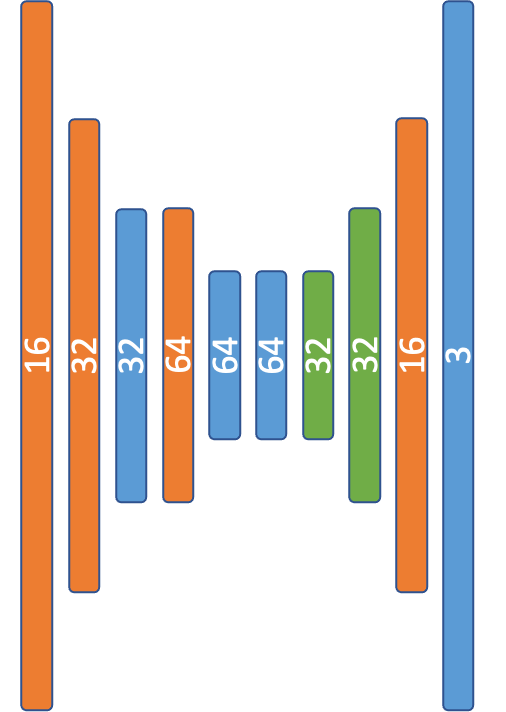}\\
        \multicolumn{3}{c}{\includegraphics[width=0.7\textwidth]{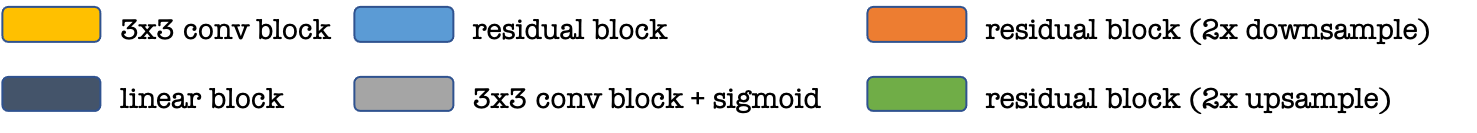}}
    \end{tabular}
    \caption{The architectures of the Global Motion Analysis (GMA) module blocks. The number of output channels is indicated in the center of each block. GAP stands for Global Average Pooling.}
    \label{fig:gma}
\end{figure}

\noindent\textbf{GMA.}
The architecture of our Global Motion Analysis (GMA) module is depicted in Fig.~\ref{fig:gma}. GMA consists of 4 networks: the masking network, 2 identical shift predictors and the inpainter. 

\begin{figure}[t]
    \centering
    \includegraphics[width=0.9\textwidth]{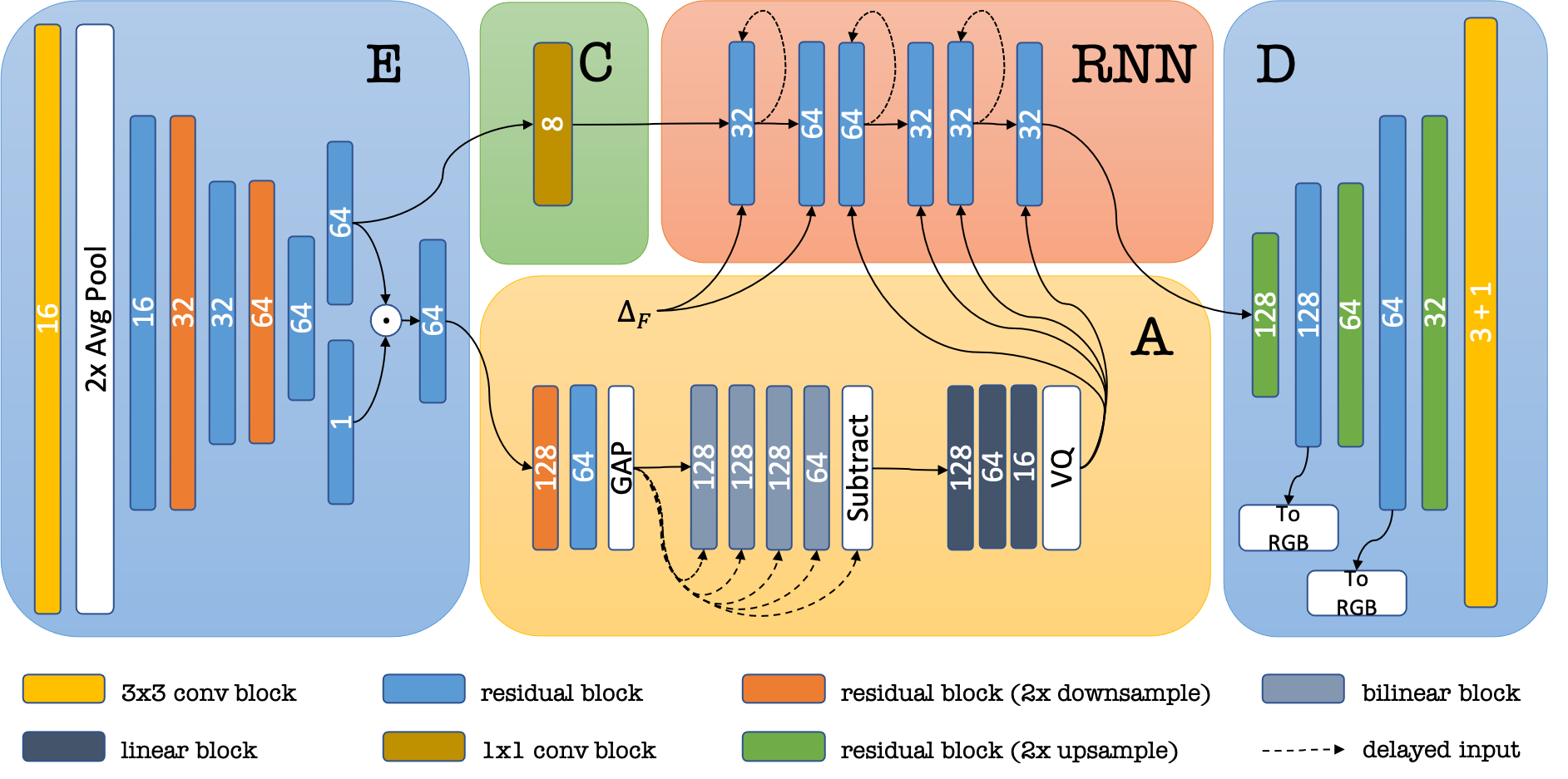}
    \caption{The architecture of the Local Motion Analysis (LMA) module of GLASS.}
    \label{fig:lma}
\end{figure}

\noindent\textbf{LMA.}
The architecture of our Local Motion Analysis (LMA) module is depicted in Fig.~\ref{fig:lma}. The encoder $E$ and the decoder $D$ are mostly adopted from Menapace et al.~\cite{menapace2021playable}. However, we introduce an additional $1\times1$-convolutional block $C$ to compress the feature vector before feeding it to the RNN. This is supposed to prevent overfitting to the appearance of the agent. We also change the RNN to take the action codes as input through the modulated convolution, as in StyleGAN \cite{karras2021alias}. Moreover, we upgrade the architecture of the action network $A$ by incorporating delayed bilinear blocks and using Vector Quantization \cite{kim2019unsupervised} for estimation of the performed action. We would also like to clarify the intuition behind using a sequence of bilinear transformations to model actions instead of the difference between $\psi_{t+1}$ and $\psi_{t}$, as done in \cite{menapace2021playable}. By using the difference as an action direction, the model only discriminates linear transitions in the latent space. This, in addition to the low dimensional action space used in \cite{menapace2021playable}, results in the fact that CADDY mostly discovers global 2D transformations, such as shifts. However, local actions are mostly periodic (consider an agent that rotates or walks in place). With our sequence of bilinear transformations we let the network unfold the latent space trajectories first before taking the difference between the features. Our ablation studies in the main paper suggest that this approach helps.

\subsection{Training details}

\noindent\textbf{Loss terms coefficients.}
The configuration of the $\lambda$ coefficients used in the linear combination of the separate loss terms is shown in Table~\ref{tab:coefs}. We found that this selection of $\lambda$ works well across all the datasets.

\begin{table}[t]
    \centering
        \caption{The coefficients of the loss terms used in the training of GLASS}
    \label{tab:coefs}
    \begin{tabular}{c|cccccc}
        \textbf{GMA} & $\lambda_\text{BIN}$ & $\lambda_\text{SIZE}$ & $\lambda_\text{RECF}$ & $\lambda_\text{RECB}$ & $\lambda_\text{VGG}$ & $\lambda_\text{RECJ}$\\
        \hline
        & 0.5 & 0.1 & 1.0 & 2.0 & 0.01 & 1.0
    \end{tabular} \\ \vspace{0.4cm}
    \begin{tabular}{c|cccccc}
        \textbf{LMA} & $\lambda_\text{VQ}$ & $\lambda_\text{RECU}$ & $\lambda_\text{RECS}$ & $\lambda_\text{MSK}$ & $\lambda_\text{CYC}$ & $\lambda_\text{LMA-VGG}$\\
        \hline
        & 0.25 & 1.0 & 1.0 & 0.2 & 0.2 & 1.0
    \end{tabular} 
\end{table}

\noindent\textbf{Sequence length scheduling.}
As described in the main paper, we choose a sequence length $T_f$, $0 < T_f < T$, after which the encodings of the reconstructed foregrounds are fed to the RNN. For all the datasets we start from $T_f = 5, T = 6$ and gradually decrease $T_f$ to 1 in 25000 iterations after the GMA pretraining stage has ended. On the BAIR dataset $T$ remains also constant, while on the Tennis and on the W-Sprites datasets we gradually increase $T$ from 6 to 12 in order to favor the quality of long generated sequences.

\noindent\textbf{Optimization and Batching.}
As mentioned in the main paper, the models are trained using the Adam optimizer \cite{kingma2014adam} with a learning rate equal to $0.0004$ and weight decay $10^{-6}$. We decrease the learning rate by a factor of $0.3$ after 300K iterations. On W-Sprites and Tennis we used batch size equal to $4$. However, on the BAIR dataset due to the high resolution of the frames, we had to decrease the batch size to $2$.

\section{W-Sprites dataset}\label{sec:data}

Here we describe how the \emph{W-Sprites} dataset was synthesized. In particular, we provide details on the random walk used to generate the global motion of the sprite. First, a starting point $(x_0, y_0)$ is sampled uniformly within the frame. At each step $i$, an action $\hat g_i$ is sampled uniformly from the list of available actions: \texttt{left}, \texttt{right}, \texttt{up}, \texttt{down} and \texttt{stay} (on the edges of the image the corresponding action is removed from the list). The transition probabilities are given by
\begin{align}
    & p(g_i = g_{i - 1} | g_{i - 1}) = p_{\text{inertia}} \\
    & p(g_i = \hat g_i | g_{i - 1}) = 1 - p_{\text{inertia}} \\
    & p(x_i = x_{i - 1} + s, y_i = y_{i - 1} | x_{i - 1}, y_{i - 1}, g_i = \text{``right''}) = 1 \\
    & p(x_i = x_{i - 1} - s, y_i = y_{i - 1} | x_{i - 1}, y_{i - 1}, g_i = \text{``left''}) = 1 \\
    & p(x_i = x_{i - 1}, y_i = y_{i - 1} + s | x_{i - 1}, y_{i - 1}, g_i = \text{``down''}) = 1 \\
    & p(x_i = x_{i - 1}, y_i = y_{i - 1} - s | x_{i - 1}, y_{i - 1}, g_i = \text{``up''}) = 1 \\
    & p(x_i = x_{i - 1}, y_i = y_{i - 1} | x_{i - 1}, y_{i - 1}, g_i = \text{``stay''}) = 1.
\end{align}
We set $p_\text{inertia}$ to 0.9 and $s$ to 7 pixels. The described process generates a sequence of coordinates $(x_i, y_i)$ and global actions $g_i$. The global actions are further used to animate the sprite. In case of \texttt{right}, \texttt{left}, \texttt{up} and \texttt{down} global actions the corresponding walking actions are applied. The \texttt{stay} action is animated with one of \texttt{slash left}, \texttt{slash right}, \texttt{slash front}, \texttt{spellcast left}, \texttt{spellcast right} and \texttt{spellcast front} chosen at random.

The same random walk is used to generate the background motion. For the background we set $p_\text{inertia} = 0.95$ and $s = 2$. We also restrict the maximum background motion to 25 pixels.

The code used to generate the dataset will be made publicly available on github.

\section{Additional Visual Examples}\label{sec:vis}

In this section we provide some additional qualitative evaluation of our method, that could not be included in the main paper due to the paper length limitations.

\noindent\textbf{More reconstruction and transfer examples.}
We include more examples of reconstruction and motion transfer using GLASS in this section. We start from an original video, which is decoded to a sequence of global and local actions. This sequence is used for both reconstructing the original video from the first frame and transfer the motion to a different scene. The results on the \emph{BAIR} and the \emph{Tennis} datasets are shown in Figs.~\ref{fig:bair_rt} and \ref{fig:tennis_rt}.

\noindent\textbf{Global action space.}
In the main paper we included some visualizations of the global action space on the \emph{BAIR} and \emph{Tennis} datasets. Here we provide more videos in order to reiterate the consistency of the global actions learnt by GLASS. We sequentially feed the same global shift to the model along with a fixed local action. The resulting 8 frames long videos are shown in Figs.~\ref{fig:bair_ga}, \ref{fig:tennis_ga} and \ref{fig:ws_ga}.

\begin{figure}
    \centering
    \includegraphics[width=0.4\textwidth]{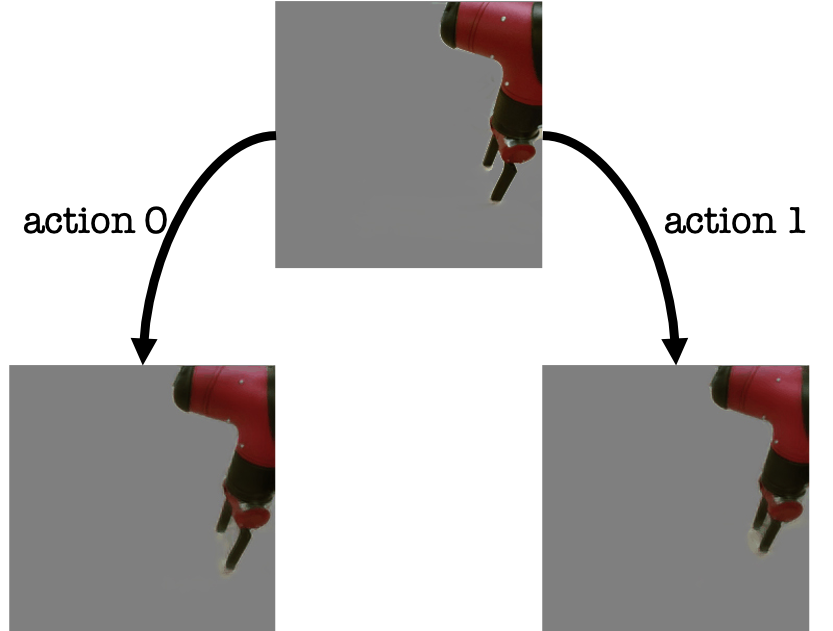}
    \caption{Demonstration of the resulting images after applying different local actions on the \emph{BAIR} dataset. The actions capture some local deformations of the robot arm, i.e. the state of the manipulator (\texttt{open} / \texttt{close}).}
    \label{fig:bair_la}
\end{figure}

\begin{figure}
    \centering
    \includegraphics[width=0.6\textwidth]{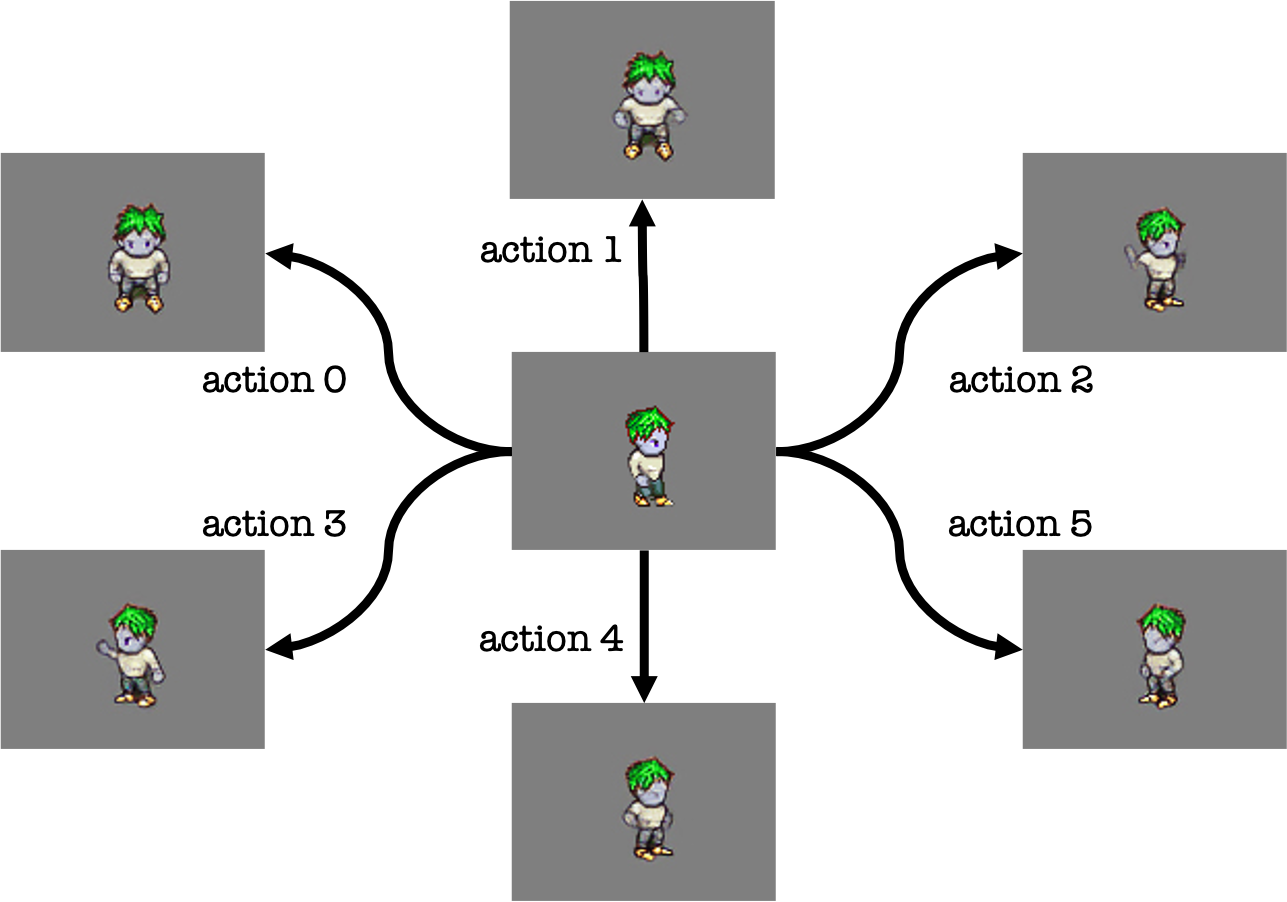}
    \caption{Demonstration of the resulting images after applying different local actions on the \emph{W-Sprites} dataset. The local actions learnt by the model can be interpreted as \texttt{turn front}, \texttt{slash front}, \texttt{spellcast}, \texttt{slash left}, \texttt{turn right}, \texttt{turn left}.}
    \label{fig:ws_la}
\end{figure}

\noindent\textbf{Local action space.}
Here we provide some visualizations of the local action space learnt by GLASS on the different datasets. In Figs.~\ref{fig:bair_la}, \ref{fig:tennis_la} and \ref{fig:ws_la} we show the first frame of the video as well as the result of applying diferent local actions. We sequentially feed the same local action to the model along with the $(0.0, 0.0)$ global action to keep the agent static. The 8th frame of the resulting sequence is shown. We fit 2, 4 and 6 local actions on the \emph{BAIR}, \emph{Tennis} and \emph{W-Sprites} datasets respectively.

\clearpage

\begin{figure}
    \centering
    \begin{tabular}{rc@{\hspace{2mm}}c@{\hspace{2mm}}c@{\hspace{2mm}}c@{\hspace{2mm}}c}
        \centered{\rotatebox[origin=c]{90}{original}} &
        \centered{\animategraphics[width=0.17\textwidth]{6}{Figures/bair_real_0/real_}{0}{11}} & 
        \centered{\animategraphics[width=0.17\textwidth]{6}{Figures/bair_real_1/real_}{0}{11}} & 
        \centered{\animategraphics[width=0.17\textwidth]{6}{Figures/bair_real_2/real_}{0}{11}} & 
        \centered{\animategraphics[width=0.17\textwidth]{6}{Figures/bair_real_3/real_}{0}{11}} & 
        \centered{\animategraphics[width=0.17\textwidth]{6}{Figures/bair_real_4/real_}{0}{11}} \\
        \centered{\rotatebox[origin=c]{90}{reconstructed}} & \centered{\animategraphics[width=0.17\textwidth]{6}{Figures/bair_rec_0/rec_}{0}{11}} & 
        \centered{\animategraphics[width=0.17\textwidth]{6}{Figures/bair_rec_1/rec_}{0}{11}} & 
        \centered{\animategraphics[width=0.17\textwidth]{6}{Figures/bair_rec_2/rec_}{0}{11}} & 
        \centered{\animategraphics[width=0.17\textwidth]{6}{Figures/bair_rec_3/rec_}{0}{11}} &
        \centered{\animategraphics[width=0.17\textwidth]{6}{Figures/bair_rec_4/rec_}{0}{11}} \\
        \centered{\rotatebox[origin=c]{90}{transfer}} &
        \centered{\animategraphics[width=0.17\textwidth]{6}{Figures/bair_tr_0/transfer_}{0}{11}} & 
        \centered{\animategraphics[width=0.17\textwidth]{6}{Figures/bair_tr_1/transfer_}{0}{11}} & 
        \centered{\animategraphics[width=0.17\textwidth]{6}{Figures/bair_tr_2/transfer_}{0}{11}} & 
        \centered{\animategraphics[width=0.17\textwidth]{6}{Figures/bair_tr_3/transfer_}{0}{11}} & 
        \centered{\animategraphics[width=0.17\textwidth]{6}{Figures/bair_tr_4/transfer_}{0}{11}}\\
    \end{tabular}
    \caption{Reconstruction and motion transfer examples on the \emph{BAIR} dataset. Note the ability of GLASS to generate very diverse videos from the same initial frame. To play use Acrobat Reader.}
    \label{fig:bair_rt}
\end{figure}

\begin{figure}
    \centering
    \begin{tabular}{rc@{\hspace{2mm}}c@{\hspace{2mm}}c}
        \centered{\rotatebox[origin=c]{90}{original}} &
        \centered{\animategraphics[width=0.3\textwidth]{6}{Figures/tennis_real_0/real_}{0}{15}} & 
        \centered{\animategraphics[width=0.3\textwidth]{6}{Figures/tennis_real_1/real_}{0}{15}} & 
        \centered{\animategraphics[width=0.3\textwidth]{6}{Figures/tennis_real_2/real_}{0}{15}} \\
        \centered{\rotatebox[origin=c]{90}{reconstructed}} & \centered{\animategraphics[width=0.3\textwidth]{6}{Figures/tennis_rec_0/rec_}{0}{15}} & 
        \centered{\animategraphics[width=0.3\textwidth]{6}{Figures/tennis_rec_1/rec_}{0}{15}} & 
        \centered{\animategraphics[width=0.3\textwidth]{6}{Figures/tennis_rec_2/rec_}{0}{15}} \\
        \centered{\rotatebox[origin=c]{90}{transfer}} & \centered{\animategraphics[width=0.3\textwidth]{6}{Figures/tennis_tr_0/transfer_}{0}{15}} & 
        \centered{\animategraphics[width=0.3\textwidth]{6}{Figures/tennis_tr_1/transfer_}{0}{15}} & 
        \centered{\animategraphics[width=0.3\textwidth]{6}{Figures/tennis_tr_2/transfer_}{0}{15}} \\
    \end{tabular}
    \caption{Reconstruction and motion transfer examples on the \emph{Tennis} dataset. Note the ability of GLASS to generate very diverse videos from the same initial frame. To play use Acrobat Reader.}
    \label{fig:tennis_rt}
\end{figure}

\clearpage

\begin{figure}[t]
    \centering
    \animategraphics[width=0.7\textwidth]{6}{Figures/bair_ga/bair_ga_}{0}{6}
    \caption{Global action space visualization on the \emph{BAIR} dataset. Each row starts with the same frame. Each column corresponds to one of the global actions, from left to right: \texttt{right}, \texttt{left}, \texttt{down}, \texttt{up} and \texttt{stay}. To play use Acrobat Reader.}
    \label{fig:bair_ga}
\end{figure}

\begin{figure}[t]
    \centering
    \animategraphics[width=0.7\textwidth]{6}{Figures/tennis_ga/tennis_ga_}{0}{6}
    \caption{Global action space visualization on the \emph{Tennis} dataset. Each row starts with the same frame. Each column corresponds to one of the global actions, from left to right: \texttt{right}, \texttt{left}, \texttt{down}, \texttt{up} and \texttt{stay}. To play use Acrobat Reader.}
    \label{fig:tennis_ga}
\end{figure}

\begin{figure}[t]
    \centering
    \animategraphics[width=0.7\textwidth]{6}{Figures/ws_ga/ws_ga_}{0}{6}
    \caption{Global action space visualization on the \emph{W-Sprites} dataset. Each row starts with the same frame. Each column corresponds to one of the global actions, from left to right: \texttt{right}, \texttt{left}, \texttt{down}, \texttt{up} and \texttt{stay}. To play use Acrobat Reader.}
    \label{fig:ws_ga}
\end{figure}

\clearpage

\begin{figure}
    \centering
    \includegraphics[width=0.8\textwidth]{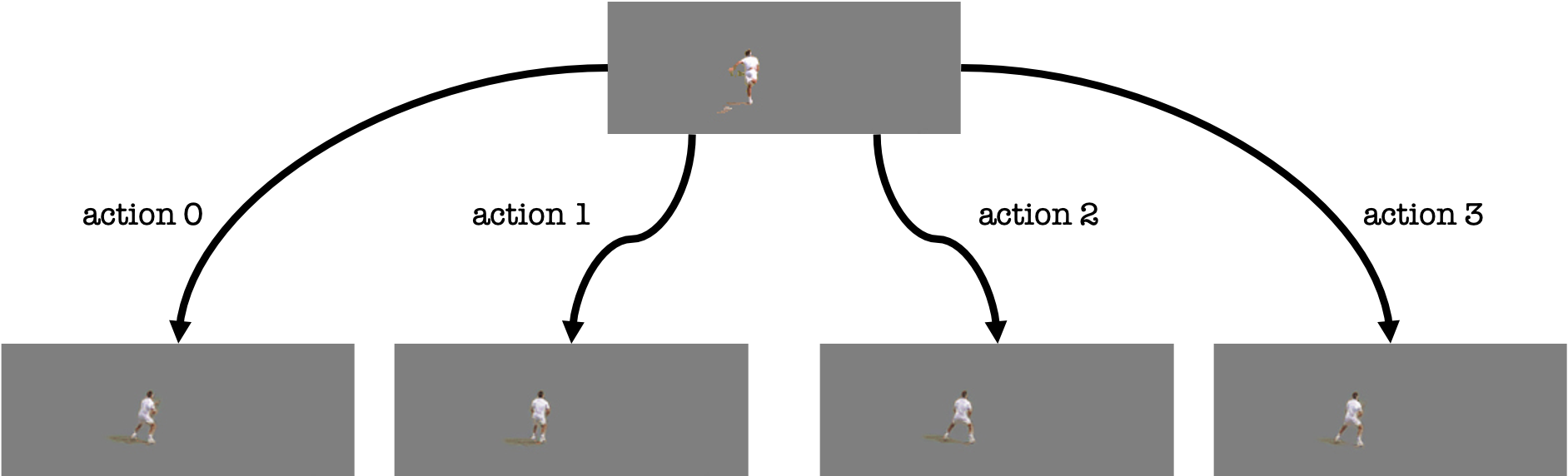}
    \caption{Demonstration of the resulting images after applying different local actions on the \emph{Tennis} dataset. The actions capture some small variations of the pose of the tennis player, such as rotation and the distance between the legs. This helps GLASS generate more realistic motions than CADDY and other competitors, e.g. running player (see Fig.~\ref{fig:tennis_rt})}
    \label{fig:tennis_la}
\end{figure}

\end{document}